\documentclass[acmsmall]{acmart}
\usepackage{graphicx}
\usepackage{float}
\usepackage{mathtools}
\usepackage{commath}
\usepackage[export]{adjustbox}
\usepackage[utf8]{inputenc} 
\usepackage[T1]{fontenc}    
\usepackage{hyperref}       
\usepackage{url}    
\usepackage{adjustbox}
\usepackage{booktabs}       
\usepackage{amsfonts}       
\usepackage{nicefrac}       
\usepackage{microtype}      
\usepackage{lipsum}
\usepackage{caption}
\usepackage{amsmath}
\usepackage{amssymb}
\usepackage{amsthm}
\usepackage{graphicx}
\usepackage[normalem]{ulem}
\useunder{\uline}{\ul}{}
\usepackage[colorinlistoftodos]{todonotes}
\usepackage{verbatim}
\usepackage[edges]{forest}
\usetikzlibrary{arrows.meta}

\usepackage{comment}

\begin{document}
\title{A Survey of Black-Box Adversarial Attacks on Computer Vision Models}

\author{Siddhant Bhambri}
\authornote{Both authors contributed equally.}
\email{siddhantbhambri_bt2k16@dtu.ac.in}
\author{Sumanyu Muku}
\authornotemark[1]
\email{sumanyumuku_bt2k16@dtu.ac.in}
\affiliation{\institution{Delhi Technological University}
    \city{Delhi}
    \country{India}}

\author{Avinash Tulasi}
\email{avinasht@iiitd.ac.in}
\author{Arun Balaji Buduru}
\email{arunb@iiitd.ac.in}
\affiliation{\institution{Indraprastha Institute of Information Technology}
    \city{Delhi}
    \country{India}
}

\renewcommand{\shortauthors}{Bhambri \& Muku, et al.}

\begin{abstract}
Machine learning has seen tremendous advances in the past few years, which has lead to deep learning models being deployed in varied applications of day-to-day life. Attacks on such models using perturbations, particularly in real-life scenarios, pose a severe challenge to their applicability, pushing research into the direction which aims to enhance the robustness of these models. After the introduction of these perturbations by Szegedy et al. \cite{Szegedy2013}, significant amount of research has focused on the reliability of such models, primarily in two aspects - white-box, where the adversary has access to the targeted model and related parameters; and the black-box, which resembles a real-life scenario with the adversary having almost no knowledge of the model to be attacked. To provide a comprehensive security cover, it is essential to identify, study, and build defenses against such attacks. Hence, in this paper, we propose to present a comprehensive comparative study of various black-box adversarial attacks and defense techniques.  
\end{abstract}




\maketitle

\section{Introduction}
\label{section_1}

Machine Learning (ML) is the most influential field in the sciences today. With applications in Computer Science, Biomedical Engineering, Finance, and Law leveraging the edge in Language, Vision, Audio with specialized branches like Deep Learning (DL), Reinforcement Learning (RL) the involvement of ML in day-to-day life is undeniable. RL models are beating world champions at their own game; DL models can classify a multitude of categories with precise identification of species, objects, material, which is humanly impossible. The field has grown so much that Deep Language models are writing fiction, news reports, and poetry at the same time. While the developments in the area are exciting, the underlying techniques are probabilistic, making the performance of an ML model highly dependent on the training data. 

Gaps in the training data have lead to biased models, a biased model can have excellent performance on a set of inputs closer to the training data, but it fails on corner cases where the input was rarely seen while training. Whatever be the domain, an ML model draws boundaries between classes in an $n$ dimensional space, and these decision boundaries can not be reasoned. Sometimes these boundaries can be subtle, making the model categorize inputs into different classes (with high confidence scores) in the immediate neighborhood of a boundary. Taking advantage of these traits in ML, to fool the system into wrongly interpreting an input, Adversarial attacks are crafted. 

A typical adversarial attack starts with studying the model, inputs, outputs, training data, model parameters, etc.  according to the threat model. While earlier threat models were surrounded around the training data and complete access to the model parameters, it is less practical in the sense that the attacker can be an outsider. The black-box adversarial attack, on the other hand, limits the attacker's capabilities, with access only to the deployed model, sometimes restrictions are also put on the observable parameters of the model. The setting of black-box attacks and constraints on attacker are closer to the real-world scenario.

To understand the Adversarial attacks in the context of Black-box in this section, we introduce the Taxonomy. 
\subsection{Taxonomy}
\label{taxo}
Here we introduce the key concepts in the context of the adversarial attacks. 
\begin{itemize}
    \item \textbf{Adversary: }An entity trying to attack the ML model to classify a legitimate-looking input wrongly. Say input $i \in c$, where c is the original class, the Adversary's goal is to make the model predict an input $i` \in c`$, where $i`$ is the input adversary carefully crafts and the point to note is $i`$ will still be classified as $c$ for a human annotator. 
    
    \item \textbf{Perturbation: }To achieve the above mentioned goals, the adversary crafts an input $i`$ which is perturbed from $i$ by adding an impurity say $\delta \dot \epsilon$. Modified input $i` = i + \delta \dot \epsilon$ is called the perturbed input. Constraints on the perturbation are a challenge to the Adversary. Any changes made should be under a certain value, to make sure that the input is still classified into its original class $c$ by a human annotator. $L2$ distance is often used in literature to define the acceptable impurity.
    \item \textbf{Adversarial Input: }Input crafted by Adversary after perturbation is called Adversarial Input. The key properties of this input are the impurity and actual input. The addition of this impurity makes the actual input "jump the classification boundary," hence the model misclassifies an Adversarial Input. 
    \item \textbf{Targeted Attack: }When the target misclassified class $c`$ of a perturbed input $i`$ is defined, then it is a Targeted Attack. Here, the adversary targets the impurity in such a way that the model classifies given input $i$ to class $c`$. 
    \item \textbf{Query: }A query is counted as a single instance of sending input to the model under attack and noting the observations made. Minimizing the number of queries made reduces the time taken to build an adversarial example, hence it is a key aspect of building efficient adversary models. 
    \item \textbf{Threat Model: }
    Threat model defines the rules of the attack, what resources can the adversary use, and the end goal of the attack. Two main components are described below:
    \begin{itemize}
        \item \textbf{Attacker Goals: }Goals define what the adversarial input seeks from the attack. Various security goals such as integrity attack, availability attack, targeted attack, exploratory attack come under goals. An attacker can attempt a single goal or multiple at a given time. 
        \item \textbf{Attacker Capabilities: }Information at adversary's disposal like the training data (with or without labels), model parameters, number of queries adversary can make come under capabilities of the attacker. Also, training time attack and post-training attack are other aspects in which the attacker's capabilities are defined. 
    \end{itemize}
    \item \textbf{White-box Attack: }In a white-box attack, the adversary knows everything about the target model. The knowledge of the adversary includes the learned weights, parameters used to tune the model. Labeled training data is also available in some cases. With this information, the usual strategy of the attacker is to model the distribution from weights and derive perturbed inputs to violate the boundaries while being in the $L2$ limits. 
    \item \textbf{Black-box Attack: }Contrast to white-box attacks, a black-box attack has limited knowledge of the model, with no labeled data sometimes. An attack with black-box constraints is often modeled around querying the model on inputs, observing the labels or confidence scores. 
    \item \textbf{Transferability: }It refers to the occasion of using a model trained on one the same of data, by observing inputs and outputs (querying) of the original model called "substitute model," and using the perturbed inputs from this model to attack the original "target model." The assumption here is that the substitute model simulates the target model. 
    
\end{itemize}

\subsection{Related Surveys - How is our survey different?}
{
    In an early work \cite{barreno2010security}, the authors introduced the now-standard taxonomy of Adversarial attacks in Machine Learning; the survey gave a structure to the interaction between attacker and defender. By fitting the existing adversarial attacks into this threat model, authors generalized on the principles involved in the security of Machine Learning. Another significant contribution of the work is a line of defenses to Adversarial attacks that the authors presented based on their taxonomy. The work \cite{gilmer2018motivating} also generalized and standardized the job done in this domain.  
    
    As the adversarial attacks became more frequent, later surveys in the field concentrated in specifics of the domain. \cite{akhtar2018threat} is an extensive survey on attacks in Computer Vision with Deep Learning, \cite{liu2018survey} gave a data-driven view of the same, and \cite{kumar2017survey} studied the fundamental security policy violations in adversarial attacks. 
    
    As mentioned in Section \ref{section_1}, Black Box adversarial attacks are more close to real life, and a lot of work is being carried out in this domain. However, extensive and informative the surveys above have been, there is a lack of study exclusive to Black Box attacks. With this work, we want to survey the existing benchmark Black Box attacks on different domains. We also address how each of the areas like Computer Vision, NLP differ in terms of robustness towards attack and the challenges in approaching these models. Table \ref{chronological_table} shows a list of works we discuss in this paper, discussing the milestone contributions in the Black Box Adversarial space.

}

\subsection{Why Black-Box Attacks?}
{
    While black-box attacks limit the capabilities of the attacker, they are more practical. In a general sense, security attacks are carried out on full-grown and deployed systems. Attacks with intent to circumvent a system, disable a system, compromise the integrity are seen in the real world. 
    
    Consideration of two parties, adversary and challenger, is possible in this paradigm. Challenger being the party that trains a model and deploys and adversary being the attacker whose intention is to break the system for a predefined goal. Multiple capability configurations that reflect real-world behavior are possible in this setting. As an example, if a robber evading from the crime scene wants to fool the surveillance system into tagging a different car number, or a criminal targeting the prey's car number getting tagged are configuration changes to adversarial user goals. However, the system under attack in this example is owned by a third party. Neither the robber nor the criminal was part of the development of the surveillance system. 
    
    The limited capabilities of the adversary are just a reflection of real-world scenarios. Hence, works in this domain are more practical. 
}

\subsection{Evaluation of Attack}
{
    Adversarial attacks, in general, are evaluated based on the number of queries made on the model to converge on parameters. Lesser the number of queries better will be the attack, as the time taken to start an attack will be minimized. Apart from the time taken to craft an attack, the perturbation norm is a standard measure of determining the effectiveness of an attack. Borrowed from Machine Learning literature, the most common perturbation norms in use are $l_2$ and $l_\infty$, which are standard error measures. 
    
    \begin{itemize}
        \item \textbf{$l_2$ norm: }Also called as Euclidean norm, $l_2$ is the shortest distance between two vectors. In the adversarial setting, the distance from the original input $i$ and the perturbed input $i`$ is calculated. To maintain the distance at minimum, the goal is to change an input $i$ in a way indistinguishable by a human annotator. 
        \item \textbf{$l_\infty$ norm: }It is the highest entry in the vector space. $l_\infty$ essentially determines the maximum value in the vector given. 
    \end{itemize}
}

\section{Black-Box vs White-Box}

{
    As discussed in Section \ref{taxo}, resources with an adversary differ considerably based on the type of attack model. In a white-box attack, an adversary can have full access to the model right from the start of training, which leads to the availability of training data, testing data, network architecture, parameters, and finally, the learned weights of a model, etc. Also, the number of queries an adversary can make comes under resources. For example, in a one-shot adversarial attack, the attacker has only a single shot at fooling the model; in such cases crafting the adversarial example does not involve a fine-tuning step. 
    
    When it comes to black-box attacks, the resources with adversary are considerably less. For one, the adversary will have no access to the model in the training phase, nor the adversary knows weights and parameters used in the model. However, once the model is deployed, based on the type of attack, the probability of each predicted label will be provided to the adversary. A stricter case is where only the predicted label is known to the attacker without any confidence score. Varying degree of information like labeled dataset is provided to adversaries in specific threat models. Again, as said above, the number of queries an attacker can make while crafting an adversarial example is considered under resources. An attack is deemed to be superior if the amount of resources consumed is minimal. 
}

\begin{table}[]
\centering
\begin{tabular}{|ccc|c|}
\hline
\textbf{Month} & \textbf{Year} & \textbf{Method} & \textbf{Proposed Approach} \\ \hline
Feb & 2016 & Papernot et al. \cite{papernot} & Local substitute model \\
Dec & 2016 & Narodytska et al. \cite{narodytska_2016} & Gradient-free \\ \hline
Jul & 2017 & Narodytska et al. \cite{narodytska_2017} & Advanced Local Search \\
Nov & 2017 & Chen et al. \cite{chen_2017} & ZOO \\ \hline
Feb & 2018 & Brendel et al. \cite{brendel} & Boundary Attack \\
Jul & 2018 & Ilyas et al. \cite{ilyas_2018} & Limited Queries and Info \\
Jul & 2018 & Cheng et al. \cite{cheng} & Opt-Attack \\
Oct & 2018 & Bhagoji et al. \cite{bhagoji} & Query Reduction using Finite Differences \\
Oct & 2018 & Du et al. \cite{du} & Attack using Gray Images \\ \hline
Jan & 2019 & Tu et al. \cite{tu} & AutoZOOM \\
Apr & 2019 & Shi et al. \cite{shi} & Curls \& Whey \\
Apr & 2019 & Dong et al. \cite{dong} & Translation-Invariant \\
May & 2019 & Chen et al. \cite{chen_2019} & POBA-GA \\
May & 2019 & Brunner et al. \cite{brunner} & Biased Sampling \\
May & 2019 & Li et al. \cite{li} & NATTACK \\
May & 2019 & Moon et al. \cite{moon} & Combinatorial Optimization \\
May & 2019 & Ilyas et al. \cite{ilyas_2019} & Bandits \&Priors \\
Jul & 2019 & Alzantot et al. \cite{alzantot} & GenAttacks \\
Aug & 2019 & Guo et al. \cite{guo} & SimBA \\ \hline
\end{tabular}%
\caption{A comparison of works that have focused on black-box scenarios in an adversarial setting.}
\label{chronological_table}
\end{table}

\section{Crafting a Black-Box attack }
{
   
A Black Box adversarial example starts with evaluating access to the resources at the attacker's disposal. These include the kind of model, any model related parameters such as confidence scores, test set, training dataset, etc. Once the resources are in place. An attack is carried out in one of the following ways. Using a \textbf{transferable} attack strategy, the adversary can choose to train a parallel model called a \textbf{substitute model} to emulate the original model. The attacker can use a much superior architecture than the original for the weight estimation. However, in the absence of a substitute model, the attacker chooses to have a \textbf{Query Feedback mechanism}, where attacker continuously crafts the perturbed input while querying on the model under attack.

 As mentioned earlier, when a network's architecture and model weights are unknown, which is the case with black-box adversarial attacks, it is common to practice for attackers to train another model from scratch, called a substitute model intending to emulate the model under attack. Once the substitute model with an attacker has achieved satisfactory accuracy, by examining the weights, adversarial examples are crafted. As the perturbed inputs start misclassified by the model at the attacker's hand, these inputs are used on the actual model to achieve the adversarial goals. The choice for a substitute model is usually kept at a much superior level to understand the latent space with superior contrast. As the adversary has much control and knowledge of the task at hand, this is the choice for one-shot and targeted adversarial attacks. To put it in a sentence: transferring the learned weights by continuous querying, and strategically attacking the model comes under transferable attacks. 
 
 Adversarial attacks without using a substitute model pose the challenge of identifying the boundaries, crafting the perturbed input at a manual level. Rather than training a model and then examining the weights, this kind of attack is more hands-on and simultaneous. The attack takes a query feedback form, where the attacker starts with random input and starts adding noise to the input under the acceptable perturbation error level. As the confidence score of input deteriorates, the attacker further pushes the noise in the direction of the noise, more like following the gradient. In some cases, gradient descent itself is used to mark the direction and track the movement of noise in the right direction. Also termed as local search, the technique boils down to searching the correct dimension in the latent space to get a misclassified input. A detailed discussion of attacks following both the substitute model and query feedback mechanism has been the pillar of black-box adversarial attacks. Beginning with a classification flowchart, the following section discusses the same in greater detail. 
 
 }

\section{Black-Box Attacks in Computer Vision}

This section provides a detailed discussion of recent and essential works in the black-box adversarial attacks domain. In this section, we cover the dataset used, key contributions, loss function used to calculate perturbations, and other relevant learnings from each of the discussed research works. 

\subsubsection{Popular Image data sets}
{
    The most frequently used datasets to demonstrate adversarial attacks are as follows: 
    \begin{itemize}
        
        \item  \textbf{MNIST} \cite{Lecun98gradient-basedlearning}: The dataset includes grayscale handwritten digits of dimensions 28x28. The dataset consists of 60,000 training examples and 10,000 test examples.
    
        \item \textbf{CIFAR 10} \cite{Krizhevsky_cifar}: The dataset includes 60,000 RGB images of dimensions 32x32. There are ten classes and 6000 images per class. The train test division consists of 50,000 training images and 10,000 test images.
        
        \item \textbf{ImageNet} \cite{imagenet_cvpr09}: ImageNet is a large scale dataset consisting of 14,197,122 images. It is described using synsets similar to WordNet. There are more than 20,000 categories that have been annotated using more than 100,000 synset categories from WordNet.
    
    \end{itemize}
}

\begin{figure}
\label{attack_classification}

        \begin{center}
        
            \begin{forest}
                for tree={
                    myleaf/.style={label={[align=left]below:{\strut#1}}},
                    s sep=1cm
                }
                [Black-Box Adversarial Attacks,rectangle,rounded corners,draw
                    [Gradient Estimation,rectangle,rounded corners,draw,align=center,
                        myleaf={$\bullet$ Chen et al. \cite{chen_2017} \\
                                $\bullet$ Ilyas et al. \cite{ilyas_2018} \\
                                $\bullet$ Cheng et al. \cite{cheng} \\
                                $\bullet$ Bhagoji et al. \cite{bhagoji} \\
                                $\bullet$ Du et al. \cite{du} \\
                                $\bullet$ Tu et al. \cite{tu} \\
                                $\bullet$ Ilyas et al. \cite{ilyas_2019}}
                    ]
                    [Transferability,rectangle,rounded corners,draw,align=center,
                        myleaf={$\bullet$ Papernot et al. \cite{papernot} \\
                                $\bullet$ Shi et al. \cite{shi} \\
                                $\bullet$ Dong et al. \cite{dong}}
                    ]
                    [Local Search,rectangle,rounded corners,draw,align=center,
                        myleaf={$\bullet$ Narodytska et al. \cite{narodytska_2016} \\
                                $\bullet$ Narodytska et al. \cite{narodytska_2017} \\
                                $\bullet$ Brendel et al. \cite{brendel} \\
                                $\bullet$ Chen et al. \cite{chen_2019} \\
                                $\bullet$ Brunner et al. \cite{brunner} \\
                                $\bullet$ Li et al. \cite{li} \\
                                $\bullet$ Alzantot et al. \cite{alzantot} \\
                                $\bullet$ Guo et al. \cite{guo}}
                    ]
                    [Combinatorics,rectangle,rounded corners,draw,align=center,
                        myleaf={$\bullet$ Moon et al. \cite{moon}}
                    ]
                ]
                \node[above=30pt,align=center,anchor=center] {Classification of prior work on black-box scenario-based attacks.};
            \end{forest}
        \end{center}
    \end{figure}     
    
\subsection{Attack Techniques}


\subsubsection{Gradient Estimation}

\begin{itemize}

    \item \textbf{ZOO}: Chen et al. \cite{chen_2017} proposed \textit{Zeroth Order Optimization(ZOO)} to estimate the gradients of target DNN in order to produce an adversarial image. The threat model assumed by the authors is that the target model can only be queried to obtain the probability scores of all the classes. The loss function formulated is:
    \begin{equation}
        For Targeted Attack: f(x,t) = max\{max_{i\neq t}log[F(x)]_i-log[F(x)]_t,-\textit{k}\}
    \end{equation}
    \begin{equation}
        For Untargeted Attack: f(x) = \{log[F(x)]_{t_0}- max_{i\neq t_0}log[F(x)]_i,-\textit{k}\}
    \end{equation}
    
    The author then uses \textit{symmetric difference quotient} method to estimate the gradient $ \frac{\partial f(x)}{\partial x_i}$:
    \begin{equation}
        \hat{g} :=  \frac{\partial f(x)}{\partial x_i} \approx \frac{f(x+he_i)-f(x-he_i)}{2h}
    \end{equation}
    
    The above naive solution requires querying the model $2p$ times, where $p$ is the dimension of the input.
    So the author proposes two \textit{stochastic coordinate methods}: ZOO-Adam and ZOO-Newton in which a gradient is estimated for a random coordinate and the update formula is obtained using ADAM \cite{adam} and Newton's Method until it reaches convergence. The authors also discuss the generation of noise in a lower dimension to improve efficiency and specify its advantages and disadvantages.
    
    \vspace{5mm}
    
    \item \textbf{Limited Queries \& Information}: The authors in \cite{ilyas_2018} take three primary cases into account to devise successful black-box adversarial attacks. The first case talks about the constraint of limited queries that can be made to the model by the adversary. This could again be of two types- one that contains a time limit and the other that is concerned with the monetary limit. The authors present a variant of Natural Evolution Strategies(NES) coupled with Projected Gradient Descent (PGD), as used in white-box attacks, to construct adversarial examples.

    The second case is modeled around the constraint of a partial-information setting where the adversary can obtain confidence scores of the first $k$ classes for a given input. These scores may not add up to 1 since the adversary doesn't have access to the probabilities for each possible classification label. To tackle this scenario, instead of beginning with the input image $x$, it is recommended to begin with $x_o$ belonging to the target adversarial class $y_{adv}$. Hence, after each step, we need to ensure two things-
    \begin{itemize}
        \item The target adversarial class needs to remain in the top-k classes at all points in time when the input image is being perturbed.
        \begin{equation}
            \epsilon_t = min\, \epsilon'\, s.t.\, rank\, \big( y_{adv}|\prod_{\epsilon'}(x^{(t-1)})\big) < k 
        \end{equation}
        \item The probability of the input image getting classified as the target class increases with every iteration of PGD.
        \begin{equation}
            x^{(t)} = arg\, \smash{\displaystyle\max_{x'}}\, P\big(y_{adv}|\prod_{\epsilon_{t-1}}(x')\big) 
        \end{equation}
    \end{itemize}
    
    Lastly, in the third scenario, the adversary is not given any confidence scores. Instead, the adversary can only obtain the names of the classification labels for the given input data. The authors define a discretized score $R(x^{t})$ for an adversarial example to quantitatively represent the adversarial nature of the input image at each step $t$, with access only to the $top-k$ sorted labels. 

    The proposed approach was evaluated on the InceptionV3 network with 78\% $top-1$ accuracy and also on Google Cloud Vision (GCV) API, which presents a real-world scenario to perform an adversarial attack. 90\% and above accuracy is achieved for all three settings when the InceptionV3 network is attacked in the techniques mentioned above.
    
    \vspace{5mm}
    
    \item \textbf{Opt-Attack}: Cheng et al. \cite{cheng} attempt to devise a black box adversarial attack in a much stricter hard-label setting. This means that querying the target model gives only the target label, unlike other threat models where $K$-class probability scores or labels are considered.
    The authors make the attack query-efficient by treating the problem as a real-valued, continuous optimization problem. 

    The objective function proposed by them is as follows:
    \begin{equation}
        Untargeted Attack: \hspace{0.2cm}
        g({\theta})= min_{\lambda>0}\hspace{0.1cm} \lambda \hspace{0.2cm}s.t\hspace{0.2cm} f({x_0+ \lambda*\frac{\theta}{||\theta||}})\neq y_0
     \end{equation}
    \begin{equation}
        Targeted Attack (given target t): \hspace{0.2cm}
        g({\theta})= min_{\lambda>0}\hspace{0.1cm} \lambda \hspace{0.2cm}s.t\hspace{0.2cm} f({x_0+ \lambda*\frac{\theta}{||\theta||}})= t
    \end{equation}
    
    Here $\theta$ is the search direction and $g(\theta)$ is the distance from the input image $x_0$ to the closest adversarial example in direction $\theta$. Chen et. al. uses ${\theta}^* = argmin_{\theta} \hspace{0.1cm}g(\theta)$ to obtain $x^* = x_0 + g({\theta}^*)*\frac{\theta}{||\theta||}$ by using a coarse grained search to initially find a decision boundary and then fine tuning the solution using Binary Search.
    The optimization problem proposed above is solved using \textit{Randomized Gradient Free (RGF) Method}, which is an iterative Zeroth Order Optimization(ZOO) Method in which the small increment value $\textbf{\textit{u}}_t$ is sampled randomly from a Gaussian Distribution.

    \vspace{5mm}
    
    \item \textbf{Query Reduction using Finite Differences}: Finite Differences method was used to estimate the gradient in the case of black-box attacks where no knowledge regarding the gradient of the loss function exists. The authors in \cite{bhagoji} propose a two-sided gradient estimation for a function $f(x)$, where $x \in R^d$.

    The loss function used in this scenario is based on logit, which was proved to work well in the white-box attacks \cite{robustness}. The logit loss is given by:
    \begin{equation}
        l(x,y) = \phi(x + \delta)_{y}\, -\, max\,\{\phi(x+\delta)_i\, : i \neq y\}
    \end{equation}
    where $y$ represents the actual label for a given sample $x$ and $\phi(.)$ are the logits.
    By taking the logarithm of the SoftMax probabilities, the logit values can be computed up to an additive constant. This additive constant gets canceled out because the difference in the logits is equal to the loss function. Hence, the proposed method of using Finite Differences can be used to calculate the difference between the logit values of the actual label $y$ and the next most likely label $y'$.

    A drawback to this naïve approach is that the model is required to make $O(d)$ queries per input, where $d$ is the dimensions of input. Authors have further proposed two query reduction techniques based on grouping the features. The first approach is based on a random selection of features and grouping them, which will allow for simultaneous gradient estimation for the selected features. The second approach is more technical since it involves computing directional derivatives along with the principal components as determined by principal component analysis (PCA).

    The proposed approach was evaluated on two data sets - MNIST and CIFAR-10. Two models, one with two convolutional layers attached to a fully connected layer and another with three convolutional layers. These models were trained on the MNIST dataset. On the other hand, ResNet-32 and ResNet-28-10 models were used for training in the case of the CIFAR-10 data set. 

    The conclusion obtained from this approach is that the effectiveness of the crafted adversarial examples can be reduced by modifying the model output probabilities on which Gradient Estimation attacks are hugely dependent on.
    
    \vspace{5mm}

    \item \textbf{Attack using Gray Images}: A completely new approach was followed in \cite{du} to generate black-box adversarial examples by initializing the adversarial sample from a gray image. This is followed by the fact that the adversary doesn't have access to any natural input. This methodology begins by initializing the input for the target model with normalized pixel values within [0,1]. Hence, the adversary now has more flexibility in adding perturbation to the image by either increasing the pixel value to 1 to make it brighter or decreasing the pixel value to 0 to make it darker.

    Then the authors have defined the fitness function in this approach as:
    
    \begin{equation}
        J(\theta) = P(y'|\theta)\approx[F(\theta)]_{y'}
    \end{equation}

    The objective here is to maximize the probability term given in the fitness function until the input image gets classified into another class y'. Similar to many other approaches, the authors have preferred to adopt NES optimization. Here $\theta$ represents the perturbed input, and Gaussian search distribution has been chosen given the fact that it provides a more straightforward method of estimating the gradient where the two parameters that need to be optimized are mean $\phi_{d}$ and variance $\sigma_{dxd}$.

    An important point to be noted in this approach is that this is a region-based attack algorithm, which implies that the gradients are dependent on the region space. Evaluation of gradients is then followed by performing a projected gradient ascent similar to the approach in \cite{ilyas_2018, towards_resistant}. Input update can be shown as:
    
    \begin{equation}
        x^t = \prod_{[0,1]}(x^{t-1}+\lambda sign(g_t(h,w)))
    \end{equation}

    Where gamma represents the learning rate, and $\pi_{[0,1]}$ represents the projection operator, which clips the values of the pixel in the range [0,1]. The search gradients are used to update the image iteratively until the goal is achieved.
    
    \vspace{5mm}
    
    \item \textbf{AutoZOOM}: Tu et al. \cite{tu} proposed the framework AutoZOOM, which is short for Autoencoder based Zeroth Order Optimization, for crafting efficient Black Box Attacks. The two main contributions of this approach were:
    (i) an adaptive random gradient estimation method for stabilizing query counts and distortion and (ii)
    an autoencoder trained offline on unlabelled data or a simple Bilinear Image Resizer(BiLIN) to 
    expedite the attack process. 
    
    The optimization problem formulated by Tu et al. \cite{tu}:
    
    \begin{equation}
        min_{x \in {[0,1]}^d} Dist(x,x_0) + \lambda. Loss(x,M(F(x)),t)
    \end{equation}
    
    where $Dist(x,x_0)={||x-x_0||}_p$ , $F(x):{[0,1]}^d \rightarrow \mathcal{R}^K$, $M(.)$ is the monotonic function applied on classifier outputs, $Loss(.)$ is the attack objective reflecting the likelihood of predicting $t = argmax_{k \in \{1,2,...K\}} [M(F(x))]_k$ and $\lambda$ is the regularization constant.
    Loss function can be both either training loss of DNN or C\&W Loss \cite{cw}.
    
    Inspired by the convergence of \textit{Symmetric Difference Quotient Method} \cite{calculus} and taking query efficiency into consideration, the authors proposed a scaled random full gradient estimator of $\nabla f(x)$, defined as:
    \begin{equation}
        \textbf{g} = b. \frac{f(x+\beta u)-f(x)}{\beta}. u
    \end{equation}
    To control error in gradient estimation, an average is taken over q random directions:
    \begin{equation}
        \bar{\textbf{g}} = \frac{1}{q} \sum_{j=1}^{q} \textbf{g}_j
    \end{equation}
    They have also proven that $l_2$ distance between $\bar{g}$ and $\nabla f(x)$ is constrained by an upper bound:
    \begin{equation}
     \begin{split}
         \mathcal{E}||\bar{g}-\nabla f(x)||_2^2 \leq & 4(\frac{b^2}{d^2}+\frac{b^2}{dq}+\frac{(b-d)^2}{d^2})||\nabla f(x)||_2^2 + \frac{2q+1}{q}b^2 \beta^2 L^2
     \end{split}   
    \end{equation}
    The above upper bound is minimum when $b \approx q$. They also mention that the convergence rate of zero order gradient descent methods is $O(\sqrt{\frac{d}{T}})$ \cite{NesterovS17, liu, pmlr-v84-wang18e} and so it is plausible to generate perturbation from a lower dimension. In AutoZOOM, random gradient estimation is performed from a lower dimension $d^{\prime}<d$ in order to improve query efficiency. A decoder,$D:\mathcal{R}^{d^{\prime}} \rightarrow \mathcal{R}^d$ is used to add noise such that: $x = x_0 + D(\delta^{\prime})$ where $\delta^{\prime} \in \mathcal{R}^{d^{\prime}}$. Initially $q=1$ but with each iteration q is increased in order to get more accurate estimation.
    
    \vspace{5mm}
    
    \item \textbf{Bandits \& Priors}: Ilyas et al. \cite{ilyas_2019} successfully exploited prior information about the gradient using bandit optimization. Two key observations can meet the challenge of obtaining prior information about the gradient:-
    
    \begin{itemize}
        \item The input data point for which gradient computation is carried out is not arbitrary, and its structure is predictable and also reflected in the gradient.
        \item Heavy correlation can be found among the gradients that are used in successive iterations while performing iterative gradient attacks (e.g., PGD).
    \end{itemize}

    Two categories of priors, namely- time-dependent priors and data-dependent priors, have been introduced, which significantly helped in performing improved black-box adversarial attacks. The value of cosine similarity at each optimization step between successive gradients proves the correlation between them. Hence, the authors used the gradient at time $t-1$ as a prior for the gradient calculation at time $t$.
    \begin{equation}
        \frac{\big \langle \bigtriangledown_{x}L(x_t,y), \bigtriangledown_x L(x_{t+1},y) \big \rangle}{\norm{\bigtriangledown_x L(x_t, y)}_2 \norm{\bigtriangledown_x L(x_{t+1},y)}_2}\quad
        t \in \{1...T-1\}
    \end{equation}

    On the other hand, an analogy to the correlation between pixel values of an image has been made to understand the relationship between gradients for two close coordinates. Hence, the structure of the input data point can also be used to reduce the query complexity.

    Untargeted adversarial examples have been crafted by considering both $l_2$ and $l_{\infty}$ norm-based threat models on the ImageNet data set. The approach experimented on three classifiers-Inception-v3, ResNet-50 and VGG16. An average of 1117 queries was required in the case of the proposed approach that made use of bandit optimization along with time and data-dependent priors as compared to about 1735 queries in the simple case of NES. Attack failure rate was decreased from 22.2\% in $l_2$ and 34\% in $l_{\infty}$ for NES to 4.6\% in $l_2$ and 15.5\% in $L_\infty$ for the Bandit approach.
    
\end{itemize}

\subsubsection{Transferability}

\begin{itemize}
    
    \item \textbf{Local Substitute Model}: In \cite{papernot}, the targeted deep learning model $O$, also known as the oracle, was used to construct a synthetic data set. Labels for this synthetically crafted input data were obtained by querying the oracle $O$ itself. This assists the attacker to create an approximation $F$ of the target model $O$. The concept of transferability allows the attacker to ensure that $O$ will also misclassify the input images misclassified by the approximated model $F$. The primary challenge faced by the attacker in this approach is to train the model $F$ without any prior knowledge of the architecture $O$. Constraint on the number of times the attacker can query the model $O$ was a limitation as well. To overcome this problem, the authors propose Jacobian-based Dataset Augmentation, which helps in approximating the decision boundaries with limited queries.
    
    Authors proposed a five-step approach, as listed below:
    \begin{enumerate}
        \item The attacker collects a small set $S_o$, which represents the entire set of inputs $S$ to the oracle. For example, a limited number of input images, say $k$ from every class will be obtained.
        \item Then, a suitable architecture is decided upon that can be used to train the approximation model $F$. It is shown that the number, size, and type of layers do not determine a successful attack.
        \item The substitute model is trained by repetitively applying the following steps-
        \begin{enumerate}
            \item Model $F$ queries the oracle model $O$ for each of the samples $\Vec{x} \in S_p$, where $S_p$ is the training set available to the substitute the model. 
            \item Approximated model is then trained by the attacker using the available training set $S_p$.
            \item Apply the augmentation technique on the training set $S_p$ available with the model $F$ and use it to generate new synthetic data points. This results in an even larger training data set $S_{p+1}$, which is a better representative of the decision boundaries of the target model.
        \end{enumerate}
        The augmentation technique proposed, i.e., the Jacobian-based Dataset Augmentation is formulated as:
        \begin{equation}
            S_{p+1} = \{\Vec{x}+\lambda\cdot sgn(J_F[\check{O}(\Vec{x})]:\Vec{x} \in S_p) \} \cup S_p 
        \end{equation}
        where, $\lambda$ is the step-size taken during augmentation to create the set $S_{P+1}$ from the initial available training set $S_p$.
    \end{enumerate}
    
    \vspace{5mm}
    
    \item \textbf{Curls \& Whey}: The authors in \cite{shi} aim to enhance the diversity as well as transferability of adversarial examples in the case of black-box attacks. Their approach consists of two steps. The first step proposed deals with the problem of adding noises/perturbation to an input sample monotonically along the gradient ascent's direction using Curl's iteration. In contrast, the second step deals with the issue of removing excessive redundant noises from the crafted adversarial examples using Whey optimization. The proposed methodology can be better understood through \cite{shi}.

    The methodology aims to push the input data point $x$ from the current class to a different class with which it shares its decision boundary. To perform the same operation, noise may be added to this data point monotonically along the direction of gradient ascent. However, the authors propose a shortcut to reach a nearer point that is present across the decision boundary. They suggest to initially follow the direction of the gradient descent and reach the 'valley floor' and then perform iterations to climb up the gradient. This is known as the Curl's iteration. Noise refinement mechanisms are followed after and before the Curl's iteration, respectively, which has been defined as Whey optimization.
    
    
    \vspace{5mm}
    
    \item \textbf{Translation-Invariant}: This attack \cite{dong} method makes use of an ensemble of images rather than aiming to optimize the objective function. The calculation of the gradient makes use of the assumption that convolutional neural networks possess the translation-invariant property. Hence, the authors propose to shift the original input image by a limited number of pixels (10, in this case) in a 2-dimensional space.
    
    The proposed method can be integrated with any of the gradient-based attacks which include Fast Gradient Sign Method (FGSM) \cite{FGSM}, Basic Iterative Method (BIM) \cite{bim}, Momentum Iterative Fast Gradient Sign Method (MI-FGSM) \cite{mifgsm}, Diverse Inputs Method \cite{diverse} and Carlini \& Wagner's Method (C\&W) \cite{cw}. Experiments have been shown by the integration of the translation-invariant property with the FGSM attack where the update rule can be stated as:
    
    \begin{equation}
        x^{adv} = x^{real} + \epsilon\dot sign(W * \bigtriangledown_{x} J(x^{real}, y))
    \end{equation}
    
    Where the notations have the usual meanings as described in \cite{FGSM}.
    
\end{itemize}

\subsubsection{Local Search}

\begin{itemize}

    \item \textbf{Gradient-free} Authors in \cite{narodytska_2016} propose an adversarial attack methodology based on the concept of greedy local-search technique. Their proposed algorithm perturbs an image by choosing the most relevant pixels of the image to which noise is added. The observed change by adding this noise in each iteration is used to approximate the loss function's gradient. The importance of the pixels can be understood in this step by observing the change in classification accuracy after each round.
    
    The most important contribution of this work is the creation of the neighboring images to a given input image. This neighborhood will consist of all images that are different from the previous round's image by one pixel only. Initially, a pixel location is randomly selected, and perturbation is added to it. The next pixel location is chosen among all the pixels that lie within a square whose side length is $2p$, and the center is the previous pixel location, where $p$ is an arbitrary parameter.
    
    The authors have focused on $k-misclassification$, where the correct label for an image should not be enlisted in the top-k classes predicted by the model. As soon as the target model can push the right class out of the top-k classes, no more noise is added to the image, and the algorithm gets terminated. The local search technique adopted is modified to increase the classification score of the targeted class in each round, in the case of targeted attacks.
    
    \vspace{5mm}
    
    \item \textbf{Advanced Local Search}: Local search technique is a commonly used incomplete search algorithm used to solve problems on combinatorics. In \cite{narodytska_2017}, the authors have used this strategy over the image space to construct an approximated network gradient. The objective function is to minimize the probability that the perturbed image does not share the classification label with the original image. The cost function defined is-
    
    \begin{equation}
        f_{c(I)}(\hat{I})=o_{c(I)}
    \end{equation}
    where $NN(\hat{I})=(o_1,...,o_C)$, in which NN is the target neural network model and $o_j$ represents the probability of image $\hat{I}$ belonging to class $j$. This function is minimized by the local search technique.
    
    The next step in the formulation is to create a set of images that can act as neighbors of the original image. The authors have defined these neighbor images with the help of pixel locations since images generated after a single iteration differ only in one pixel from the previous image. Consider the pixels that were perturbed in the last round. Hence, in the current round, all those pixels that lie in a square of side length $2d$ with the perturbed pixel at its center will be considered. Here, $d$ is only a parameter.
    
    In the last step, a function $g$ has been defined as the transformation function, which generates a new image $\hat{I}_i$ after every round $i$ by adding perturbation to a set of pixels only.
    
    \vspace{5mm}
    
    \item \textbf{Boundary Attack}: Decision-based \cite{brendel} adversarial attacks that are based on a model's final decision, i.e., class labels or transcribed sentences, are more common in real-life scenarios as compared to confidence score-based attacks. The authors also argue that these attacks are more robust to defense techniques such as gradient masking, intrinsic stochasticity, or vigorous training. Also, lesser information regarding the targeted model is required to conduct a successful attack. This was the first time that decision-based attacks were introduced that focus on deep learning models using real-life data sets such as ImageNet.
    
    The proposed algorithm initializes by taking a random distribution of pixels for an image in the range of [0, 255]. Using this image, the algorithm traverses the decision boundary of the class to which the sample input image belongs to. The initialized image is then able to reduce its distance w.r.t. the input image besides staying inside the adversarial region. 
    
    \vspace{5mm}

    \item \textbf{POBA-GA}: Chen et al. \cite{chen_2019} proposed three parameters to be used to generate different perturbations- different noise point pixel thresholds, number of noise points, and noise point size. The next step is to add this noise to the first adversarial example to generate the next generation of the population. For evaluation purposes, a fitness function $\phi$ is defined, followed by checking if the adversary has satisfied the termination condition or not. Lastly, to generate the next set of the population for optimizing the perturbation, the suggested approach follows the typical operations of a genetic algorithm, i.e., selection, crossover, and mutation.
    
    The importance of the proposed fitness function is that it helps to evaluate how useful the adversarial input is at a given iteration. Hence, the convergence of the model largely depends on this function. The authors have incorporated two critical qualities of adversarial images into the definition of the proposed function which are
    
    \begin{equation}
        \phi(AS^t_i) = P(AS^t_i) - \frac{\alpha}{max Z(A^0)}Z(A^t_i)
    \end{equation}

    where the left-hand side of the equation represents the fitness function for Adversarial Sample $i$, $(AS_i)$ is the probability term which accounts for the effectiveness of the attack for the sample to get misclassified and $Z(A^t_i)$ is a metric originally proposed in this methodology to account for the perturbation introduced. $\alpha$ is used as a parameter to control the proportion for the fitness function between attack performance and the perturbation size. To reduce the number of queries to the target model, alpha is initially taken as zero and later assigned a value when the attack is successful. The fitness function finally looks like-
    
    \begin{equation}
        \phi (AS^t_i) = 
        \begin{cases}
            {p(y_1|AS^t_i) - p(y_0|AS^t_i) - \frac{\alpha}{max Z(A^{t_0})}Z(A^t_i)} \quad \quad {y_1 \neq y_0}\\
            {p(y_2|AS^t_i) - p(y_0|AS^t_i)}\quad \quad \quad \quad \quad \quad \quad \quad \quad \, \, {y_1 = y_0}
        \end{cases}
    \end{equation}
    
    \vspace{5mm}
    
    \item \textbf{Biased Sampling}: The authors in \cite{brunner} adopt a unique approach to introduce perturbations for creating adversarial samples known as Perlin Noise \cite{perlin}. This method was originally used to generate procedural texture patterns for computer graphics but can also be used to develop parametrized patterns having low-frequency. In the case of an original Boundary Attack \cite{brendel}, an orthogonal perturbation $\eta^k$ is applied along with a hyper-spherical surface present around the original image. However,  Perlin noise distribution is used to create a sample to introduce a low-frequency before the Boundary Attack. Usually, a permutation vector $v$ with a size of 256 is used to parametrize the noise. The noise patterns are sampled in two directions- along with the height and width of the image due to which the patters get strongly concentrated in the directions of low-frequency.
    
    Image masking has been performed to calculate the difference between the pixel values of the original image and the adversarial image and can be shown as:
    
    \begin{equation}
        M = \abs {X_{adv} - X_{orig}}
    \end{equation}

    This calculation is performed after every step and applied on every single element of the orthogonal perturbation $\eta^k$ that was sampled previously.
    
    \begin{equation}
        \eta^k_{biased} = M \odot \eta^k\, ;\, \eta^k_{biased} = \frac{\eta^k_{biased}}{\norm{\eta^k_{biased}}}
    \end{equation}

    Hence, the distortion of only the pixels with a considerable difference value is increased, and it remains the same for the remaining pixels, further reducing the search space effectively.
    
    \vspace{5mm}
    
    \item \textbf{$\mathcal{N}$-Attack}: A black box adversarial attack has been proposed in \cite{li}, which tries to learn the probability density over a small region $(l_p \hspace{0.1cm}ball)$ centered around the input such that a sample drawn from the density is likely an adversarial image. Their work is similar to Ilyas et al. \cite{ilyas_2018} who have employed the loss function: $f(x^{\prime}) := -F(x^{\prime})_y $. They have argued that Ilyas et al.'s \cite{ilyas_2018} algorithm is inferior as it relies on accurate estimation of gradients, which is not possible when DNN is non-smooth. For estimating gradients $\nabla f(x_t)$ with derivative-free methods the authors have also used \textit{NES} \cite{NES}.
    
    The smooth optimization criteria proposed by the authors is:
    \begin{equation}
    min_{\theta}J(\theta):= \int f(x^{\prime}) \pi_S(x^{\prime}|\theta) dx^{\prime}
    \end{equation}
    
    where $f(x^{\prime})$ is the C\&W Loss \cite{cw}, $\pi_S(x^{\prime}|\theta)$ is the probability density in region S $(l_p \hspace{0.1cm} ball)$ i.e $||x^{\prime}-x||_p \leq \epsilon$ and also $dim(\theta) << dim(x)$. The authors reformulated the problem with change of variables:
    \begin{equation}
        x^{\prime} = proj_S(g(z)),\hspace{0.3cm} z \sim \mathcal{N}(z|\mu,\sigma^2)
    \end{equation}
    where $g:\mathbb{R}^{dim(\mu)} \mapsto \mathbb{R}^{dim(x)}$. The transformation steps involve sampling z
    i.e $z \sim \mathcal{N}(\mu,\sigma^2)$, then $g(z)$ is computed: $g(z)= \frac{1}{2(tanh(g_0(z))+1)}$. After
    this, $\delta^{\prime}=clip_p(g(z)-x)$ for p = 2 or $\infty$:
    \begin{equation}
        clip_2(\delta) = 
        \begin{cases}
            \delta\tau_2 / ||\delta||_2 & if ||\delta||_2 > \tau_2 \\
            \delta & else
        \end{cases}
    \end{equation}
    \begin{equation}
        clip_{\infty} = min(\delta, \tau_\infty)
    \end{equation}
    
    Updated $J(\theta)$ due to variable transformation:
    \begin{equation*}
        J(\theta)= \mathbb{E}_{\mathcal{N}(z|\mu,\sigma^2)}f(proj(g(z)))
    \end{equation*}
    
    $\theta=(\mu,\sigma^2)$ are the unknowns. $\sigma$ is found using Grid Search and \textit{NES} \cite{NES}
    is used to find $\mu$:
    
     \begin{equation*}
     \mu_{t+1} \leftarrow \mu_t - \eta \nabla_{\mu}J(\theta)|_{\mu_t}
     \end{equation*}
   
    \begin{equation}
        \mu_{t+1} \leftarrow \mu_t -\frac{\eta}{b} \sum_{i=1}^{b}f(proj_S(g(z_i)))\nabla_{\mu}log \mathcal{N}
      (z_i|\mu_t,\sigma^2)
    \end{equation}
    
    The process is iteratively repeated until arriving at an sampled image $x^{\prime}$ such that $C(x^{\prime}) \neq C(x)$.
    
    \vspace{5mm}
    
    \item \textbf{GenAttack}: Alzantot et al. \cite{alzantot} use genetic algorithms, which are population-based gradient-free optimization strategies to produce an adversarial image. The author justifies using the above approach by mentioning the overhead produced by excessive querying in \textit{Gradient Estimation Methods} and also the added advantage of being able to bypass defenses based on Gradient Masking and Obfuscation. The Genetic Algorithm is based on the process of natural selection, which involves iteratively choosing the \textit{\textbf{P}} set of candidates that constitute a \textit{generation} in each iteration. The quality of the candidates is assessed using a \textit{fitness} function, and the candidates for the next generation are obtained using \textit{crossover}, i.e., breeding of parents selected randomly to produce a child and \textit{mutation} which is the noise added to the child to reflect the diversity of the population. The fitness function proposed here can be shown as:
    \begin{equation}
        Fitness(x) = log{f(x)}_t - log{\sum_{j=0,j\neq t}^{j=k}f(x)_c}
    \end{equation}
    Here $f: \mathbb{R}^d->[0,1]^K $ is the black box classifier which outputs probabilities of the $K$ classes and $t$ is the label for targeted attack. The parents for crossover are randomly sampled using $probs$, i.e. the probabilities which are obtained using softmax of the fitness values. Crossover operation involves selecting features of $parent_1$ and $parent_2$ with $(p,1-p)$ respectively where $p = \frac{fitness(parent_1)}{fitness(parent_1)+ fitness(parent_2)}$. A Child is mutated using the following formula $child_{mut} = child+ \mathit{Bernoulli}(p)* \mathfrak{\textbf{U}}(-\alpha \delta_{max},\alpha\delta_{max})$. $\alpha,\rho$ and $\delta_{max}$ are mutation range, mutation probability and $l_{\infty}$ metric respectively.
    
    \vspace{5mm}
    
    \item \textbf{SimBA}: The study shown in \cite{guo} proposes a straightforward yet efficient approach to generate adversarial examples in a black-box setting. From a set $Q$ of orthonormal vectors, the algorithm picks any single vector $q$, and the modification made to the input image $x$ can be expressed as $x + q\epsilon$ or $x - q\epsilon$. The idea behind this is to modify and add noise to the image in a particular direction and check if this brings about a change in the probability $P(y|x)$, which represents the probability of the label being $y$ given the input image $x$. Here, $\epsilon$ represents the step size and can be controlled by the adversary. To keep the number of queries to the target model minimum, the methodology proposed ensures that when the algorithm picks $q \in Q$, there are no two directions that cancel out each other.
    
    Inspired by the work of Guo \cite{low_frequency}, the authors suggest making use of low-frequency noise to be added to the input image for generating better adversarial examples. They employ the discrete cosine transform (DCT) to map signals in a 2D image space $R^{dxd}$ to frequency coefficients belongs to cosine wave function magnitudes.
    
\end{itemize}

\subsubsection{Combinatorics}

\begin{itemize}

     \item \textbf{Combinatorial Optimization}: In the approach presented in \cite{moon}, the optimization problem has been treated as a discrete surrogate problem. The authors support this my mentioning that first-order attacks like FGSM \cite{FGSM} and PGD \cite{PGD} have a good performance due to their constrained nature and also due to querying the gradient of the loss concerning the input.
    Using the Taylor Expansion, the loss and the optimization problem becomes:
    \begin{equation*}
        l(x_{adv},y) \approx l(x,y) + (x_{adv}-x)^T \nabla_x l(x,y)
    \end{equation*}
    \begin{equation}\label{parsi_eq}
         maximize_{||x_{adv}-x||_{\infty} \leq \epsilon} \Rightarrow maximize_{x_{adv}} {x_{adv}}^T \nabla_x
        l(x,y) \hspace{0.3cm} subject \hspace{0.1cm} to \hspace{0.1cm} - \epsilon \textbf{1} \preceq x_{adv}-x \preceq \epsilon \textbf{1} 
    \end{equation}
    
    Due to the bounds present in the equation \ref{parsi_eq}, the authors concluded that the solution of the above linear program would be obtained at the vertex of the $l_{\infty}$ ball. So the reformulated optimization problem becomes:
    \begin{equation}
        maximize_{\textit{S} \subseteq \textit{V}} \left\{ F(\textit{S}) \triangleq f \left(x + \epsilon \sum_{i \in \textit{S}} e_i -\epsilon \sum_{i \notin \textit{S}} e_i \right)    \right\}
    \end{equation}
    
    where $f(x)=l(x,y_{gt})$ for untargeted attacks and $f(x) = -l(x,y_{target})$ for targeted attacks. They then explain the property of submodularity, a function $F:2^{\textit{V}} \rightarrow \mathcal{R}$ is submodular if for every $\textit{A} \subseteq \textit{B} \subseteq \textit{V}$ and $e \in \textit{V} \setminus \textit{B}$: $\Delta (e|\textit{A}) \geq \Delta(e| \textit{B}) $. So submodular functions exhibit a diminishing property as set size increases. With submodularity and using greedy style algorithms, it is possible to obtain $(1-\frac{1}{e})$ approximation of global optimum for monotone submodular functions and $\frac{1}{3}$ approximation for non-monotone submodular functions. The authors use a modified version of local search algorithm by Fiege et al. \cite{Feige} which alternates between greedily inserting the element while the marginal gain is strictly positive $\left( \Delta\left(e|\textit{S} > 0  \right) \right) $ and removing the element while the marginal gain is also strictly positive.
    
\end{itemize}

\subsection{Comparative Analysis of Black Box attack techniques}

The attack methods proposed have been tested quantitatively on three primary datasets, as mentioned earlier, i.e., MNIST, CIFAR-10, and ImageNet. Therefore, we present a qualitative comparison between the results of these techniques based. The results can be first divided broadly into two categories: Targeted \& Untargeted attacks for each of the three datasets. In each table, we mention the model used for generating the adversarial attack, the average number of queries made to the targeted model, average $L_2$ distance between the original image and the perturbed image, attack success rate and lastly some specific parameters mentioned in the experiments of the respective study. The results for different comparative parameters are obtained from standard performance metrics as produced by the individual research studies.

\begin{table}[htbp]
  \centering
  \begin{adjustbox}{width = \textwidth, center}
    \begin{tabular}{|c|c|c|c|c|c|}
    \hline
    \textbf{Method} & \textbf{Model} & \textbf{Average \# Queries} & \textbf{Average $L_2$} & \textbf{Attack Success Rate} & \textbf{Specific Parameters Used} \\
    \hline
    Chen et al. \cite{chen_2017}     & InceptionV3 & -     & 1.19916 & 88.90\% & Step Size = 0.002, \# Iterations = 1500 \\
    \hline
    Brendel et al. \cite{brendel}     & ResNet-50 & 192102 & 4.876 & 100\% & - \\
    \hline
    Cheng et al. \cite{cheng}     & ResNet-50 & 145176 & 4.934 & 100\% & - \\
    \hline
    Moon et al. \cite{moon}    & InceptionV3 & 722   & -     & 98.50\% & $L_{\infty}$ setting, $\epsilon$ = 0.05, Max \# Queries = 10,000 \\
    \hline
    Guo et al. \cite{guo}    & InceptionV3 & 1283  & 3.06  & 97.80\% & Step Size = 0.2, \# Iterations = 10,000 \\
    \hline
    \end{tabular}%
    \end{adjustbox}
  \caption{Comparison of methods used for generating an untargeted attack on the ImageNet data set.}
  \label{imagenet_untargeted}%
\end{table}%

From Table \ref{imagenet_untargeted}, we can see that the attack success rate for \cite{brendel} and \cite{cheng} is 100\%. However, the technique proposed by \cite{moon} takes the least number of queries to produce a successful black-box attack. 

\begin{table}[htbp]
  \centering
    \begin{adjustbox}{width = \textwidth, center}
    \begin{tabular}{|c|c|c|c|c|c|}
    \hline
    \textbf{Method} & \textbf{Model} & \textbf{Average \# Queries} & \textbf{Average $L_2$} & \textbf{Attack Success Rate} & \textbf{Specific Parameters Used} \\
    \hline
    Chen et al. \cite{chen_2017}     & InceptionV3 & -     & 3.425 & - & Batch Size = 128, \# Iterations = 20,000 \\
    \hline
    Ilyas et al. \cite{ilyas_2018}     & InceptionV3 & 11550 & -     & 99.20\% & $L_{\infty}$ setting, $\epsilon$ = 0.05 \\
    \hline
    Du et al. \cite{du}     & InceptionV3 & 1701  & 24.323 & 100\% & Image size-64x64, Momentum factor = 0.4 \\
    \hline
    Tu et al. \cite{tu}    & InceptionV3 & 13525 & 0.000136 & 100\% & Regularization coefficient = 10 \\
    \hline
    Chen et al. \cite{chen_2019}    & InceptionV3 & 3573  & 4.559451 & 95\%  & \# Iterations = 400 \\
          & ResNet-50 & 3614  & 3.0575142 & 98\%  & \# Iterations = 400 \\
          & VGG19 & 3786  & 4.023045 & 96\%  & \# Iterations = 400 \\
    \hline
    Brunner et al. \cite{brunner}    & InceptionV3 & -     & -     & 85\%  & Max \# Queries = 15,000, $L_2$ setting, $\epsilon$ = 25.89  \\
    \hline
    Moon et al. \cite{moon}    & InceptionV3 & 7485  & -     & 99.90\% & $L_{\infty}$ setting, $\epsilon$ = 0.05, Max \# Queries = 100,000 \\
    \hline
    Ilyas et al. \cite{ilyas_2019}    & InceptionV3 & 1858  & -     & 84.50\% & Max \# Queries = 10,000, $L_2$ setting \\
          & InceptionV3 & 1117  & -     & 95.40\% & Max \# Queries = 10,000, $L_{\infty}$ setting \\
          & ResNet-50 & 993   & -     & 90.30\% & Max \# Queries = 10,000, $L_2$ setting \\
          & ResNet-50 & 722   & -     & 96.60\% & Max \# Queries = 10,000, $L_{\infty}$ setting \\
          & VGG16 & 594   & -     & 82.80\% & Max \# Queries = 10,000, $L_2$ setting \\
          & VGG16 & 370   & -     & 91.60\% & Max \# Queries = 10,000, $L_{\infty}$ setting \\
    \hline
    Alzantot et al. \cite{alzantot}    & InceptionV3 & 11081 & 61.68669 & 100\% & Max \# Queries = 1,000,000, $L_{\infty}$ setting, $\epsilon$ = 0.05 \\
    \hline
    Guo et al. \cite{guo}    & InceptionV3 & 7899  & 9.53  & 100\% & Step Size = 0.2, \# Iterations = 30,000 \\
    \hline
    \end{tabular}%
    \end{adjustbox}
    \caption{Comparison of methods used for generating a targeted attack on the ImageNet data set.}
  \label{imagenet_targeted}%
\end{table}%

A lot of models such as Inception-V3, ResNet-50, VGG16, and VGG19 have been used to experiment with targeted attacks on ImageNet data set, and this can be seen in Table \ref{imagenet_targeted}. The attacks mentioned in \cite{du} and \cite{alzantot} suffer significantly in terms of the average $L_2$ distance parameter as compared to the others. The authors in \cite{ilyas_2019} used three different models and experimented with their attack by keeping a limit on the max number of queries at 10,000 in $L_2$ and $L_{\infty}$ settings. The number of queries was less than 2000 and, in most cases, below 1000, whereas the attack success rate ranged from 82.80\% using the VGG16 model to 96.60\% in ResNet-50.

\begin{table}[htbp]
  \centering
  \begin{adjustbox}{width = \textwidth, center}
    \begin{tabular}{|c|c|c|c|c|c|}
    \hline
    \textbf{Method} & \textbf{Model} & \textbf{Average \# Queries} & \textbf{Average $L_2$} & \textbf{Attack Success Rate} & \textbf{Specific Parameters Used} \\
    \hline
    Chen et al. \cite{chen_2017}     & -     & -     & 1.4955 & 100\% & Batch Size = 128, Step Size = 0.01, \# Iterations = 3000 \\
    \hline
    Brendel et al. \cite{brendel}     & ResNet-50 & 131972 & 1.11545 & 100\% & - \\
    \hline
    Cheng et al. \cite{cheng}     & ResNet-50 & 67036 & 1.0827 & 100\% & - \\
    \hline
    Bhagoji et al. \cite{bhagoji}     & -     & -     & 2.7   & 99.7  & Batch Size = 100 \\
    \bottomrule
    \end{tabular}%
    \end{adjustbox}
    \caption{Comparison of methods used for generating an untargeted attack on the MNIST data set.}
  \label{mnist_untargeted}%
\end{table}%

In Table \ref{mnist_untargeted}, three out of four methods achieve a 100\% success rate, whereas only the attack proposed by \cite{cheng} surpasses the others in the remaining two parameters.

\begin{table}[htbp]
  \centering
  \begin{adjustbox}{width = \textwidth, center}
    \begin{tabular}{|c|c|c|c|c|c|}
    \hline
    \textbf{Method} & \textbf{Model} & \textbf{Average \# Queries} & \textbf{Average $L_2$} & \textbf{Attack Success Rate} & \textbf{Specific Parameters Used} \\
    \hline
    Chen et al. \cite{chen_2017}     & - & -     & 1.9871 & 98.90\% & Batch Size = 128, Step Size = 0.01, \# Iterations = 3000 \\
    \hline
    Brendel et al. \cite{brendel}     & ResNet-50 & 93543 & 2.0626 & 100\% & - \\
    \hline
    Cheng et al. \cite{cheng}     & ResNet-50 & 59094 & 1.7793 & 100\% & - \\
    \hline
    Bhagoji et al. \cite{bhagoji}     & Self-designed, accuracy = 99.2\% & 62720 & -     & 100\% & Batch Size = 100, $L_{\infty}$ setting, $\epsilon$ = 0.3 \\
    \hline
    Tu et al. \cite{tu}    & C\&W \cite{cw}  & 2428  & 3.55936 & 100\% & Regularization coefficient = 0.1 \\
          & C\&W \cite{cw}  & 730   & 3.23792 & 100\% & Regularization coefficient = 1 \\
          & C\&W \cite{cw}  & 510   & 3.66128 & 100\% & Regularization coefficient = 10 \\
    \hline
    Chen et al. \cite{chen_2019}    & - & 423   & 2.352 & 100\% & \# Iterations  = 100 \\
    \hline
    Alzantot et al. \cite{alzantot}    & -     & 996   & -     & 100\% & Step Size = 1.0, Max \# Queries = 100,000 \\
    \bottomrule
    \end{tabular}%
    \end{adjustbox}
    \caption{Comparison of methods used for generating a targeted attack on the MNIST data set.}
  \label{mnist_targeted}%
\end{table}%

Almost all the methods in Table \ref{mnist_targeted} show a 100\% attack success rate. The attacks proposed in \cite{tu, chen_2019, alzantot} achieve an attack in less than 1000 queries.

\begin{table}[h!]
  \centering
    \begin{adjustbox}{width = \textwidth, center}
    
    \begin{tabular}{|c|c|c|c|c|c|}
    \hline
    \textbf{Method} & \textbf{Model} & \textbf{Average \# Queries} & \textbf{Average $L_2$} & \textbf{Attack Success Rate} & \textbf{Specific Parameters Used} \\
    \hline
    Chen et al. \cite{chen_2017}     & -     & -     & 0.19973 & 100\% & Batch Size = 128, Step Size = 0.01, \# Iterations = 1000 \\
    \hline
    Brendel et al. \cite{brendel}     & ResNet-50 & 384204 & 4.8758 & -     & - \\
    \hline
    Cheng et al. \cite{cheng}     & ResNet-50 & 145176 & 4.9339 & -     & - \\
    \hline
    Bhagoji et al. \cite{bhagoji}     & ResNet-32 & -     & -     & 100\% & Batch Size = 100 \\
    \hline
    Moon et al. \cite{moon}    & ResNet-32 & 1261  & -     & 48\%  & Max \# Queries = 20,000 \\
    \bottomrule
    \end{tabular}%
    \end{adjustbox}
    \caption{Comparison of methods used for generating an untargeted attack on the CIFAR-10 data set.}
  \label{cifar_untargeted}%
\end{table}%

In Table \ref{cifar_untargeted}, the black-box attack proposed by \cite{chen_2017} achieves an exceptional performance by getting a 100\% success rate while getting an average $L_2$ distance of 0.19973. \cite{moon} uses a ResNet-32 model to achieve an average of 1261 queries. However, the attack success rate is as low as 48\%.

\begin{table}[htbp]
  \centering
    \begin{adjustbox}{width = \textwidth, center}
    \begin{tabular}{|c|c|c|c|c|c|}
    \hline
    \textbf{Method} & \textbf{Model} & \textbf{Average \# Queries} & \textbf{Average $L_2$} & \textbf{Attack Success Rate} & \textbf{Specific Parameters Used} \\
    \hline
    Chen et al. \cite{chen_2017}     & same as 14 & -     & 0.39879 & 96.80\% & Batch Size = 128, Step Size = 0.01, \# Iterations = 1000 \\
    \hline
    Brendel et al. \cite{brendel}     & ResNet-50 & 170972 & 0.2395 & -     & - \\
    \hline
    Cheng et al. \cite{cheng}     & ResNet-50 & 130020 & 0.2476 & -     & - \\
    \hline
    Bhagoji et al. \cite{bhagoji}     & ResNet-32 & 61440 & -     & 100\% & Batch Size = 100, $L_{\infty}$ setting, $\epsilon$ = 8 \\
    \hline
    Tu et al. \cite{tu}    & C\&W \cite{cw}  & 1524  & 3.6864 & 100\% & Regularization coefficient = 0.1 \\
          & C\&W \cite{cw}  & 332   & 0.00101 & 100\% & Regularization coefficient = 1 \\
          & C\&W \cite{cw}  & 259   & 0.00115 & 100\% & Regularization coefficient = 10 \\
    \hline
    Chen et al. \cite{chen_2019}    & same as 4 & 381   & 0.2088 & 100\% & \# Iterations  = 100 \\
    \hline
    Alzantot et al. \cite{alzantot}    &       & 804   & -     & 96.50\% & Step Size = 1.0, Max \# Queries = 100,000 \\
    \bottomrule
    \end{tabular}%
    \end{adjustbox}
    \caption{Comparison of methods used for generating a targeted attack on the CIFAR-10 data set.}
  \label{cifar_targeted}%
\end{table}%

As shown in Table \ref{cifar_targeted}, again the models proposed in \cite{tu, chen_2019, alzantot} achieve a high attack success rate percentage while using less than 1000 queries on the targeted model. Most of the attacks shown in this table perform well in terms of the average $L_2$ distance parameter.

\section{Defense against Adversarial attacks}

In this section, we provide a detailed discussion of recent works in the defenses proposed against adversarial attacks. Conventional paradigms in defenses follow gradient masking, input transformation, adversarial training, adversarial detection-based approaches. The following subsections discuss each of these paradigms, with key examples from each concept.

\begin{table}[]
\centering
\begin{tabular}{|ccc|c|}
\hline
\textbf{Month} & \textbf{Year} & \textbf{Method} & \textbf{Proposed Approach} \\ \hline
Mar & 2016 & Papernot et al. \cite{papernot_defense} & Distillation \\ \hline
Mar & 2017 & Hosseini et al. \cite{hosseini} & NULL Labeling \\
Sep & 2017 & Meng et al. \cite{meng} & MagNet \\
Dec & 2017 & Xu et al. \cite{xu} & Feature Squeezing \\ \hline
Jan & 2018 & Guo et al. \cite{guo_defense} & Input Transformations \\
Feb & 2018 & Buckman et al. \cite{buckman} & Thermometer Encoding \\
Feb & 2018 & Samangouei et al. \cite{samangouei} & Defense-GAN \\
Feb & 2018 & Xie et al. \cite{xie_2018} & Randomization \\
Mar & 2018 & Dhillon et al. \cite{dhillon} & SAP \\ 
Mar & 2018 & Kannan et al. \cite{kannan} & ALP \\
Mar & 2018 & Prakash et al. \cite{prakash} & Pixel Deflection \\
Apr & 2018 & Shaham et al. \cite{shaham} & Basis Functions Transformation \\
May & 2018 & Liao et al. \cite{liao} & Guided Denoiser \\
May & 2018 & Song et al. \cite{song} & Pixel Defend \\
Jun & 2018 & Wong et al. \cite{wong} & Convex Adversarial Polytope \\
Sep & 2018 & Tram$\acute{e}$r et al. \cite{tramer} & Ensemble Adversarial Training \\
Sep & 2018 & Madry et al. \cite{madry} & M-PGD \\
Nov & 2018 & Rakin et al. \cite{rakin} & PNI \\ \hline
Mar & 2019 & Xie et al. \cite{xie_2019} & Feature Denoising \\
Jun & 2019 & Cohen et al. \cite{cohen} & Randomized Smoothing \\
Jul & 2019 & Mustafa et al. \cite{mustafa} & Hidden Space Restriction \\  \hline
\end{tabular}%
\caption{A comparison of works that have focused on defenses against an adversarial setting.}
\label{chronological table defenses}
\end{table}

\begin{figure}
        
        \begin{center}
        
            \begin{forest}
                for tree={
                    myleaf/.style={label={[align=left]below:{\strut#1}}},
                    s sep=0.5cm
                }
                [Defense Mechanisms against Adversarial Attacks,rectangle,rounded corners,draw
                    [Gradient Masking,rectangle,rounded corners,draw,align=center,
                        myleaf={$\bullet$ Papernot et al. \cite{papernot_defense} \\
                                $\bullet$ Xie et al. \cite{xie_2018} \\
                                $\bullet$ Dhillon et al. \cite{dhillon}}
                    ]
                    [Input Transformation,rectangle,rounded corners,draw,align=center,
                        myleaf={$\bullet$ Guo et al. \cite{guo_defense} \\
                                $\bullet$ Buckman et al. \cite{buckman} \\
                                $\bullet$ Samangouei et al. \cite{samangouei} \\
                                $\bullet$ Prakash et al. \cite{prakash} \\
                                $\bullet$ Shaham et al. \cite{shaham} \\
                                $\bullet$ Liao et al. \cite{liao} \\
                                $\bullet$ Song et al. \cite{song}}
                    ]
                    [Adversarial Training,rectangle,rounded corners,draw,align=center,
                        myleaf={$\bullet$ Hosseini et al. \cite{hosseini} \\
                                $\bullet$ Tram$\acute{e}r$ et al. \cite{tramer} \\
                                $\bullet$ Kannan et al. \cite{kannan} \\
                                $\bullet$ Rakin et al. \cite{rakin} \\
                                $\bullet$ Xie et al. \cite{xie_2019} \\
                                $\bullet$ Cohen et al. \cite{cohen} \\
                                $\bullet$ Mustafa et al. \cite{mustafa} \\
                                $\bullet$ Madry et al. \cite{madry}}
                    ]
                    [Adversarial Detection,rectangle,rounded corners,draw,align=center,
                        myleaf={$\bullet$ Meng et al. \cite{meng} \\
                                $\bullet$ Xu et al. \cite{xu} \\
                                $\bullet$ Wong et al. \cite{wong}}
                    ]
                ]
                \node[above=30pt,align=center,anchor=center] {Classification of prior work on defense strategies proposed against adversarial attacks.};
            \end{forest}
        \end{center}
        \label{defense_classification}
    \end{figure}

\subsection{Defense Techniques}

    \subsubsection{Gradient Masking}
    
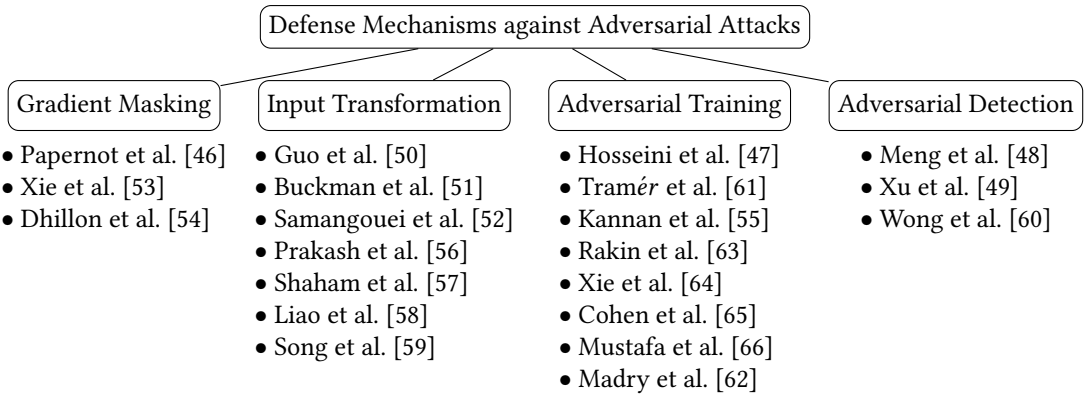
    \begin{itemize}

    \item \textbf{Distillation}: The defense mechanism proposed by the authors \cite{papernot_defense} is based on $\textit{distillation}$ introduced by Hinton et al. \cite{DBLP:journals/corr/HintonVD15}. Distillation, in an easy sense, is used to transfer knowledge from larger to smaller DNN architectures to reduce computational complexity. Given a training data set $\left\{ (X, Y(X)):X \in \mathcal{X} \right\}$ where $X$ are the images, and $Y$ are the one-hot vector(e.g., \{0,0,1...0\}) spread over the length of classes with the correct class having 1, a deep neural network $F$ is trained with softmax output layer with temperature $T$ which is used to control the knowledge transfer. $F(X)$ is the probability vector over all the classes. A second training set $\left\{ (X,F(X)):X \in \mathcal{X}\right\}$ is obtained and $F(X)$ obtained from the first DNN are here used as the \textit{soft labels}. A second model, $F^d$, is trained using this dataset, and this model is called a \textit{distilled} model. This model is much more smoothed. The amplitude of its network gradients exploited by adversarial samples are also thereby reduced.

     \item \textbf{Randomization}: The goal of this paper \cite{xie_2018} is to alleviate adversarial attacks by introducing two random layers at inference time: 1) \textit{Random Resizing} 2) \textit{Random Padding}. These randomization effects help to defend against single-step and iterative gradient-based attacks. The random resizing layer converts an image of size $W{*}H{*}3$ into a new image of size $W{'}{*}H{'}{*}3$ such that $|W{'}-W|$ and $|H{'}-H|$ are within a reasonably small range. The random padding layer pads zeros around the resized image in a random manner. Suppose after random padding the size of the image is $W{''}{*}H{''}{*}3$ then there are $(W{''}-W{'}+1){*}(H{''}-H{'}+1)$ different possible patterns.\\

     \item \textbf{SAP}: The authors \cite{dhillon} have proposed \textit{Stochastic Activation Pruning} as a method for guarding pre-trained networks against adversarial examples. During the forward pass, they stochastically prune a subset of activations in each layer. After pruning, they scale up the surviving activations to normalize the dynamic range of inputs to the subsequent layers. Deriving inspiration from game theory \cite{Game_theory}, they treat the problem as a minimax zero-sum game between adversary and the model:
    
    \begin{equation}
      \pi^{*},\rho^{*} := arg \hspace{0.1cm}min_{\pi} max_{\rho} \mathbb{E}_{p \sim \pi,r \sim \rho}[J(M_p(\theta),x+r,y)],   
    \end{equation}
    
    where $\rho$ is the adversary's policy from which $r \sim \rho$ perturbation is sampled and $\pi$ is the defender's policy from which $p \sim \pi$ model parameters corresponding to $M_p(\theta)$ is sampled. The adversary's goal is to maximize the defender's loss under strategy $\rho$ and defender's goal is to minimize the loss by changing model parameters $\theta$ to $M_p(\theta)$ under strategy $\pi$. Now given the $i$'th layer activation map,$h^i \in \mathbb{R}^{a^i}$, the probability of sampling $j$'th activation with value $(h^i)_j$ is given by:
    
    \begin{equation*}
        p_{j}^{i} = \frac{|(h^i)_j|}{\sum_{k=1}^{a^i}|(h^i)_k|}
    \end{equation*}
    
    If activation is sampled, it is scaled up by inverse of the probability of sampling over all draws. If not, then the activation is set to 0. In this way, the inverse propensity scoring of each activation is preserved. Under an instance $p$ of policy $\pi$, $r_{p}^{i}$ sample are drawn giving a new activation map $M_p(h^i)$:
    
    \begin{equation*}
        M_p(h^i) = h^i \odot m_{p}^{i} \hspace{0.5cm} (m_{p}^{i})_j = \frac{\mathbb{I}((h^i)_j)}{1-(1-p_{j}^{i})^{r_{p}^{i}}}
    \end{equation*}
    
    This scheme is similar to dropout \cite{JMLR:v15:srivastava14a}; however, it differs from it in its sampling technique: SAP is more likely to sample activations that are high in absolute value whereas dropout samples each activation with the same probability.

    \end{itemize}

    \subsubsection{Input Transformation}
    \begin{itemize}
    \item \textbf{Input Transformations}: The authors \cite{guo_defense} tested several input transformations-1) Image Cropping and Rescaling 2) Bit Depth Reduction \cite{xu} 3) JPEG Image Compression and Decompression \cite{JPEG} 4) Total Variance Minimization \cite{TV_MINI} 5) Image Quilting \cite{Image_Quilt} and their corresponding effects in withstanding different adversarial attacks. Specifically, \textit{total variance minimization} and \textit{image quilting} were effective defenses when the model is trained on transformed images. Their strength lies in their \textit{non-differential nature} and \textit{randomness} which makes it difficult for the adversary to circumvent them.
    Total Variance Minimization involves synthesizing an image $\mathbf{z}$ from adversarial input x by solving:
    \begin{equation}
      min_{z}||(1-X) \odot (z-x) ||_{2} + \lambda_{TV}.TV_p(\mathbf{z})  
    \end{equation}
    
    This approach selects a small set of pixels and reconstructs the image $x$ that is consistent with the selected pixels. The pixels are selected by sampling a Bernoulli Random Variable $X(i,j,k)$ at each pixel position $(i,j,k)$, and the pixel is set when $X(i,j,k)=1$. $\odot$ is the element-wise multiplication.
    
    \begin{equation}
        TV_p(\mathbf{z})= \sum_{k=1}^K \left[ \sum_{i=2}^N ||\mathbf{z}(i,:,k)-\mathbf{z}(i-1,:,k)||_p + \sum_{j=2}^N ||\mathbf{z}(:,j,k)- \mathbf{z}(:,j-1,k)||_p \right]
    \end{equation}
    
    $TV_p(z)$ is the total $l_p$ variation of z. The total variation(TV) measures the fine-scale variation in z, thereby encouraging the removal of adversarial perturbation during TV minimization.\\
    Image Quilting is a technique that involves synthesizing images by combining images obtained from a database. The algorithm places image patches for a predefined set of grid points and computes minimum graph cuts in overlapping boundary regions to remove edge artifacts. \newline

     \item \textbf{Thermometer Encoding}: Buckman et al. \cite{buckman} proposes a \textit{input discretization} method to break the linear extrapolation behaviour of machine learning models. Goodfellow et al. \cite{exp_harness_adv} provided evidence that several network architectures are vulnerable to adversarial examples as the loss function of these networks tend to be highly linear to its inputs. Though neural networks, in principle, can represent highly non-linear functions, networks trained via SGD tend to converge to linear solutions. One hypothesis explaining this states that the nonlinearities used in networks are piece-wise linear. Instead of using highly complex non-linear functions that are difficult to train and generalize, the authors have specified a simple non-differentiable and non-linear transformation(\textit{discretization}). Two discretization methods have been specified: 1) \textbf{One Hot Encoding}: for a pixel $\mathit{i} \in \{1,2,...n \}$:
    \begin{equation}
        f_{oneshot}(x_i) = \chi(b(x_i)) 
    \end{equation}
    where $b(\theta)$ is the quantization function. For $b_i= \frac{i}{k}$, $b(\theta)$ is the largest index $\alpha \in \{1,2,...k \}$ such that $\theta \leq b_{\alpha}$. $\chi(j)$ is the one hot vector of $j$ i.e $\chi(j) \in \mathbb{R}^k$ such that:
    
    \begin{equation*}
    \chi(j)_l =
        \begin{cases}
                 & 1 \hspace{0.3cm} if \hspace{0.1cm} l=j \\
        & 0 \hspace{0.3cm} otherwise
        \end{cases}
    \end{equation*}
    
    So for an input pixel $i$, $b(.)$ obtains the index $\alpha$ such that $i \leq b_{\alpha}$. $\chi(.)$ computes the one hot vector spread across the length of the input dimension and sets 1 at index $\alpha$ and 0 everywhere else.
    
    2) \textbf{Thermometer Encoding}: For a pixel $i \in \{1,2,...k \}$:
    \begin{equation}
        f_{therm}(x_i) = \tau(b(x_i)) = \mathbb{C}(f_{oneshot}(x_i))
    \end{equation}
    where $\mathbb{C}$ is the cumulative sum function, $\mathbb{C}(c)_l = \sum_{j=0}^l c_l$. $\tau(j)$ is the thermometer vector i.e $\tau(j) \in \mathbb{R}^k$ such that:
    
    \begin{equation*}
    \tau(j)_l = 
        \begin{cases}
        & 1 \hspace{0.3cm} if\hspace{0.1cm} l>=j \\
        & 0 \hspace{0.3cm} otherwise
        \end{cases}
    \end{equation*}
    
    $\tau(j)$ computes the thermometer vector, which is one hot vector with consecutive one after the index $\alpha$ obtained using the $b(.)$ function. \newline
    
     \item \textbf{Defense-GAN}: Samangouei et al. \cite{samangouei} proposed a defense mechanism based on \textit{Wasserstein-GAN} \cite{wgan}. A GAN is a model in which min-max game is played between two adversaries:- \textit{a generator(G)} and \textit{a discriminator(D)}. The \textit{W-GAN} is similar to \textit{GAN}, but instead of min-max loss, Wasserstein loss is used for learning the distribution of unperturbed images. At inference time, it finds a close output to the given input image which does not contain adversarial changes i.e., for an input image $x$ we obtain $z^*$ such that:
    
    \begin{equation}
        z^* \hspace{0.1cm} = \hspace{0.1cm} min_z ||G(z)-x||_{2}^{2}
    \end{equation}
    
    The above minimization problem is solved by doing L steps of Gradient Descent on R different random initializations of z. Then $G(z^*)$ is added as the input to the classifier. The motivation of the above mechanism is derived from the global optimality of \textit{GAN} min-max loss when $p_g=p_{data}$ i.e., the probability that the image is sampled from data distribution is equal to that of the generator and if \textit{G} and \textit{D} have enough capacity to represent the data, and the training algorithm is such that $p_g$ converges to $p_{data}$ then:
    
    \begin{equation}
        \mathbb{E}_{x \sim p_{data}} \left[ min_z ||G_t(z)-x||_2\right] \rightarrow 0 
    \end{equation}
    where $\mathbb{E}$ is the expected reconstruction loss which after $t$ steps of training converges to 0, thereby indicating minuscule performance loss on clean examples. This minimization of reconstruction loss helps in reducing the adversarial noise. \newline

    \item \textbf{Pixel Deflection}: The authors have proposed two methods for defending against adversarial attacks: 1) Pixel Deflection samples pixels randomly in an input image and replaces it with another randomly sampled pixel from its square neighborhood. Intuitively, Prakash et al. \cite{prakash} show that adversarial attacks tend to add perturbations in the entire plane of the image, disregarding the location of the object. The background pixels are not that useful in the classification of the object. By \textit{Pixel Deflection}, certain pixels are dropped, preserving enough background for classification and mitigate the impact of adversarial attacks. The probability for a pixel to be dropped is inversely proportional to the probability of that pixel constituting the part of an object i.e
    \begin{equation*}
        Pr_{u}(P_i) \propto \frac{1}{Pr_{o}(P_i)}
    \end{equation*}
    2) Wavelet Denoising is used to reduce noise introduced by both pixel deflection and adversarial attacks. Prior works \cite{Wavelet_denoising_1,Wave_denoising_2,Wave_den_3} illustrate certain regularities in wavelet responses of natural images that can be used to denoise images. The author's method uses \textit{Discrete Wavelet Transform} \cite{DWT}, which converts the image signal into orthonormal wavelets. These wavelets form the basis for an image's space, orientation, and scale, etc. \newline
    
    \item \textbf{Basis Function Transformations}: Shaham et al. \cite{shaham} augments images at inference time using basis functions. The transformations smoothen the images thereby reducing adversarial noise. Basis functions are known functions (usually polynomial) on which linear regression is performed in order to model a non-linear function. The basis functions used in this paper include: 1) \textit{Low Pass Filtering} 2) \textit{Principal Component Analysis} 3) \textit{Wavelet Approximation} 4) \textit{Soft-thresholding} 5) \textit{JPEG Compression} \newline
    
    \item \textbf{ Guided Denoiser}: Since adversarial samples are based on adding noise to clean images, the natural idea that came to the authors was to develop denoising models as a defense mechanism \cite{liao}. They defined two denoising networks:\\ 1)\textbf{Pixel Guided Denoiser(PGD)}: Given a clean image $x$, the denoiser $D:x{*} \rightarrow \hat{x}$ is used to convert the adversarial image $x{*}$ into $\hat{x}$ such that $L = ||x-\hat{x}||$ is minimized where L is the $L_1$ norm. Two denoising functions were discussed-1) \textit{Denoising Autoencoder(DAE)} \cite{DAE} and \textit{Denoising U-Net(DUNET)} which was DAE modified with U-Net \cite{DUNET}.

    2)\textbf{High-Level representation Guided Denoiser(HGD)}: HGD uses the same U-Net structure as DUNET but differs with it in terms of the loss function. Given a target neural network $f$, the HGD minimizes the loss $L=||f(x)_l-f(\hat{x})_l||$ where $f(x)_l$ is the representation of the neural network at layer $l$. Three variants of HGD were also specified-1) When $l=-2$, then the representations of the top-most
    convolutional layer is used, and the denoiser is called \textit{Feature Guided Denoiser}. 2)When $l=-1$,
    logit layer representations are used, and the denoiser is called \textit{Logits Guided Denoiser}. 3)When the classification loss of the target model is used as a denoising loss function, then the denoiser is called \textit{Class Guided Denoiser}. \newline

    \item \textbf{PixelDefend}: The authors \cite{song} have proposed the defense mechanism \textit{PixelDefend} which aims to detect adversarial images and modify the input images such that it resembles more the images derived from the training distribution $p(x)$. The training distribution $p(x)$ is learnt through a PixelCNN \cite{PixelCNN}(p(x)=$p_{CNN}(x)$). The probabilities of the input image, $X{'} \sim q(X)$ and training images, $\{X_1,X_2,...X_N \} \sim p(X)$ are evaluated using PixelCNN and rank of $p_{CNN}(X')$ in $\{p_{CNN}(X_1),p_{CNN}(X_2),...p_{CNN}(X_N)\} $ is used as a text-statistic:
    
    \begin{equation*}
     T = T(X{'};X_1....X_N) \triangleq \sum_{i=1}^{N} \mathbb{I}[p_{CNN}(X_i) \leq p_{CNN}(X{'})].   
    \end{equation*}
    
    The above equation is used to calculate the \textit{p-value}, which is used as a metric to decide whether an image is adversarial or not:
    
    \begin{equation}
        p = \frac{1}{N+1} \left( \sum_{i=1}^{N}\mathbb{I}[T_i \leq T] +1 \right) = \frac{T+1}{N+1} = \frac{1}{N+1} \left( \sum_{i=1}^{N} \mathbb{I}[p_{CNN}(X_i) \leq p_{CNN}(X{'})]+1 \right)
    \end{equation}
    
    Now the input image $X$ is used to calculate the image $X^*$ that maximizes $p(X)$ subject to the constraint that $X^*$ is within the $\epsilon_{defend}$ ball(maximum perturbation) of $X$:
    
   \begin{align*}
       & max_{X^*}p(X^*) \\
       s.t \hspace{0.2cm} & ||X-X^*||_{\infty} \leq \epsilon_{defend}
   \end{align*}
    
    Iteratively pixels ($ x \leftarrow X[i,j,k]$) are sampled from the range $R \leftarrow [max(x-\epsilon_{defend},0),min(x+ \epsilon_{defend},255)]$, with the goal to maximize the training data distribution $p(X^*)$.
   
    \end{itemize}

    \subsubsection{Adversarial Training}
    \begin{itemize}
        
    \item \textbf{NULL Labeling}: In this paper \cite{hosseini}, the authors propose to block transferability by training a classifier such that as input noise level increases, the classifier shows lower confidence on the original label and instead declares the input invalid. The method consists of three steps: 1) initial training, 2) computing NULL probabilities for adversarial examples generated with different amounts of perturbations 3)adversarial training.

    The compute function $f$, which is used for computing the null probability $p_{NULL}$ for adversarial examples, in this case, is taken as the \textbf{MG Method}. During the Adversarial training phase, the adversarial examples are generated using \textbf{STG Method}. The network is trained based on the following optimization problems:
   
    \begin{align*}
       & \delta X^* = argmin_{\delta X} l(X+\delta X,Z_T;\theta), \hspace{0.2cm} s.t. ||\delta X||_0 \sim U[1,N_{max}] \\
       & \theta^* = argmin_{\theta} \alpha l(X,Z;\theta) + (1-\alpha)l(X+ \delta X^*,Z_A;\theta)  
    \end{align*}
    
    Here $l(X,Z;\theta)$ is the loss function with the parameters $\theta$. $X$ and $X^*$ are the input and adversarial images, respectively. $\delta X$ is the noise sampled from the uniform distribution $U$ in the range $[1,N_{max}]$ where $N_{max}$ is the minimum number for which $f(\frac{N_{max}}{|X|})=1$ where $f$ is the classifier. $Z_T$ is the probability vector that is used for generating adversarial examples, whereas $Z_A$ is the desired output probability vector. $\alpha$ is the hyperparameter that controls the training between clean and adversarial samples. \newline

     \item \textbf{Ensemble Adversarial Training}: Tramer et al. \cite{tramer} hypothesized that Kurakin et al.'s \cite{kurakin_2017b} adversarially trained model on single-step attacks was still vulnerable to multi-step attacks. He proposed \textit{Ensemble Adversarial Training} as a method for decoupling the adversarial examples from the parameters of the trained model and increase the diversity of perturbations seen during training. The algorithm augments the training dataset of the target model with adversarial examples transferred from \textit{other static pre-trained} models. Intuitively, the transfer of adversarial examples from multiple models acts as a good approximation of maximization problem in Madry et al.'s \cite{madry} paper, which was not possible from transferable single step attacks. \newline
     
    \item \textbf{ALP}: In this paper \cite{kannan}, Kannan et al. proposed a method called as \textit{logit pairing} and its 2 variants- \textit{clean} and \textit{adversarial}. They compared their method with a modified version of defense proposed by Madry et al. \cite{madry2018} which they formulated as \textbf{mixed minibatch PGD}. \textbf{M-PGD} is stated as min-max problem:
    \begin{equation}
        argmin_{\theta} \left[ \mathbb{E}_{(x,y)\in \hat{p}_{data}} \left( max_{\delta \in \mathit{S}}L(\theta,x+\delta,y)\right)+ \mathbb{E}_{(x,y)\in \hat{p}_{data}} \left( L(\theta,x,y) \right) \right]
    \end{equation}
    \textbf{Adversarial and Clean Logit Pairing} involve minimizing following losses respectively:
    \begin{equation}
    \label{adv_logit_pair}
      J(\mathbb{M},\theta) + \lambda \frac{1}{m} \sum_{i=1}^{m} L\left( f(x^{(i)};\theta),f(\bar{x}^{(i)};\theta) \right)  
    \end{equation}
    
    \begin{equation}
    \label{clean_logit_pair}
         J^{(clean)}(\mathbb{M},\theta) + \lambda \frac{2}{m} \sum_{i=1}^{\frac{m}{2}} L\left( f(x^{(i)};\theta),f(x^{(i+\frac{m}{2})};\theta) \right)
    \end{equation}
    
    Here $J(\mathbb{M},\theta)$ is the classifier loss for minibatch $\mathbb{M}$. $\lambda$ and m are the penalizing coefficient and training size respectively. $f(x,\theta)$ is the logit layer and \textbf{L} is the loss function for calculating the similarity of two images. Here \textbf{L} is the $L^2$ distance metric and in equation [\ref{adv_logit_pair}] $(x,\bar{x})$ are the training and its adversarial image respectively and in equation [\ref{clean_logit_pair}] $(x,\Acute{x})$ are two random training images respectively. \newline
    
    \item \textbf{M-PGD}: Madry et al. \cite{madry} conducts a careful examination of the optimization problem in the study of adversarial attacks and robust models. He proposes a guarantee that a adversarial robust model should satisfy in the form of a \textit{Saddle-Point} problem:
    
    \begin{equation}
    min_{\theta} \rho(\theta),\hspace{0.4cm}\text{where}\hspace{0.2cm} \rho(\theta) = \mathbb{E}_{(x,y)\sim D}\left[max_{\delta \in \textit{S}}L(\theta,x+\delta,y) \right] 
    \end{equation}
    
    Evaluation of local maxima of loss values by using \textit{PGD} \cite{PGD} attack on \textit{MNIST} and \textit{CIFAR-10} models suggest that PGD is the universal adversary among the first-order approaches. Robustness against PGD yields robustness against all first-order adversaries i.e., as long as the adversary uses the gradients of the loss function w.r.t input, it will not find significantly better local maxima than PGD. So to maximize the inner optimization problem, PGD is the best first-order attack, whereas adversarial training is the solution for the outer optimization problem. The authors also provide a comparative study of network capacity against adversarial robustness. For a weaker \textit{black-box} attack based on transferability, increased network capacity decreases transferability. Adversarial training using stronger adversaries also reduces transferability. \newline
    
    \item \textbf{PNI}: Rakin et al. \cite{rakin} has proposed \textit{Parametric Noise Injection} which aims to make the model more robust by adding trainable gaussian noise at each layer on either activation or weights during the adversarial training phase:
    
    \begin{align}
        &\bar{v_i} = f_{PNI}(v_i) = v_i +\alpha_i.\eta; \hspace{0.4cm} \eta \sim \mathcal{N}(0,\sigma^2) \\
        &where \hspace{0.2cm} \sigma = \sqrt{\frac{1}{N}\sum_i(v_i-\mu)}
    \end{align}
    
    Here $v_i$ is the element of noise-free tensor $v$, which can be input/weight/activation. $\eta$ is the noise sampled from the gaussian distribution, and $\alpha_i$ is the learnable noise parameter, which is the same along with a layer to avoid over-parameterization and facilitate convergence. The gradient computation of loss w.r.t to noise is:
    
    \begin{equation*}
        \frac{\partial \mathcal{L}}{\partial \alpha}=\sum_i \frac{\partial \mathcal{L}}{\partial f_{PNI}(v_i)} \frac{\partial f_{PNI}(v_i)}{\partial \alpha}; \hspace{0.3cm} \frac{\partial f_{PNI}(v_i)}{\partial \alpha} = \eta
    \end{equation*}
    
    Using a gradient descent optimizer with momentum, the optimization of $\alpha$ at step $j$ can be written as:
    
    \begin{equation*}
        V_{i}^{j} = m.V_{i}^{j-1} + \frac{\partial \mathcal{L}^{j-1}}{\partial \alpha}; \hspace{0.3cm} \alpha^j = \alpha^{j-1} - \epsilon.V_{i}^j
    \end{equation*} 
    
    where m is the momentum, $\epsilon$ is the learning rate, and $V$ is the updating velocity.\newline
   
    \item \textbf{Feature Denoising}: Xie et al. \cite{xie_2019} suggests that small perturbations added in the input domain lead to substantial noise in the feature maps of the network. While the feature maps of clean images focus on semantically relevant content, the feature maps of adversarial images are even activated in the irrelevant regions.

    Motivated by this, he proposes \textit{feature denoising} as a defense mechanism. Adversarial Training of the classifier is performed together with the addition of a \textit{denoising block}. The structure of the block is inspired from \textit{self-attention} \cite{attention} and \textit{Non-Local Networks} \cite{NonLocal2018}.

    Several denoising operations have been mentioned in this paper: 1) \textbf{Non-Local Means} \cite{nonlocal-algo}: For a input feature map $x$, a denoised feature map $y$ is computed by weighted mean of features in all spatial locations $\mathcal{L}$:
    
    \begin{equation}
    \label{fd_eq}
        y_i = \frac{1}{\mathcal{C}(x)} \sum_{\forall j \in \mathcal{L}}f(x_i,x_j).x_j,
    \end{equation}
    
    where $f(x_i,x_j)$ is the feature dependent weighting function and $\mathcal{C}(x)$ is the normalization function. 
    Two Variants have been discussed- 1) \textit{Gaussian(softmax)} sets where $f(x_i,x_j)=e^{\frac{1}{\sqrt{d}}\theta(x_i)^T \phi(x_j)}$ and $\mathcal{C} = \sum_{\forall j \in \mathcal{L}}f(x_i,x_j)$ and 2) \textit{Dot Product} sets where $f(x_i,x_j)=x_{i}^{T}x_j$ and $C(x)=\mathit{N}$ where $\mathit{N}$ is the number of pixels in $x$. 2) \textbf{Bilateral Filter} \cite{Bilateral}:It is similar to Equation[\ref{fd_eq}] but differs from it in its neighborhood $\Omega(i)$, which is a local region (e.g a 3X3 patch) around pixel $i$. 3) \textbf{Mean Filter}: Simplest denoising operation is the mean filter i.e average pooling with a stride of 1. Mean filters reduce noise but also smooth structures. 4) \textbf{Median Filter}: It is defined as $y_i = median\{\forall j \in \Omega(i):x_j\}$ where the median is over a local region $\Omega(i)$ and is performed separately for each channel. It is good at removing \textit{salt and pepper} noise.\newline
    
    \item \textbf{Randomized Smoothing}: The goal of this defense method is to obtain a new classifier $g$ from base classifier $f$ which returns the most probable class when the input $x$ perturbed with Gaussian Noise is added to the base classifier:
    \begin{equation}
        g(x) = arg max_{c \in \mathcal{Y}} \hspace{1mm} \mathbb{P}(f(x+ \epsilon)=c), \hspace{2mm} \epsilon \sim \mathcal{N}(0,\sigma^2 I)
    \end{equation}
    
    Now $n$ samples of Gaussian noise are drawn and added to the input image $x$. These noisy samples of x are given as input to the base classifier $f$ and the class counts are obtained.
    If $\hat{c_A}$ and $\hat{c_B}$ are the two classes with maximum counts $n_A$ and $n_B$ then a Binomial Hypothesis test is performed to obtain the p-value ($n_A$ $\sim$ Binomial($n_A+n_B$,p)). If the p-value is less than a constant $\alpha$ then return $\hat{c_A}$ otherwise abstain from returning an output. \newline
    
    \item \textbf{ Hidden Space Restriction}: The authors of this paper suggest that the main reason for perturbations is the close-proximity of different class samples in the learned space. To counter this, they proposed to disentangle class-wise intermediate feature representation of deep networks i.e., force the features of each class to lie inside a convex region, which is maximally separated from the regions of other classes. For a input sample $x$ let the adversarial polytope (convex region within which the label of the input image doesn't change) be $P_{\epsilon}(x;\theta)={F_{\theta}(x+\delta) \hspace{0.1cm}s.t.,||\delta||_p\leq \epsilon }$, then $O_{\epsilon}^{i,j}$ represents the overlap region between data pair $(i,j)$ i.e $O_{\epsilon}^{i,j}=P_{\epsilon}(x_{y_i}^i;\theta)\cap P_{\epsilon}(x_{y_j}^j;\theta)$. The authors propose that reducing this overlap region between different classes will result in lower adversarial success. To achieve this, the authors augment the loss function used while training. This new loss added is called \textit{Prototype Conformity Loss}:
    
    \begin{equation}
      \mathcal{L}_{PC}(x,y) = \sum_{i} \left\{ ||f_i-w_{y_i}^c||_2 - \frac{1}{k-1}\sum_{j\neq y_i}\left(    ||f_i-w_{j}^{c}||_2 + ||w_{y_i}^{c}-w_{j}^{c}||_2     \right) \right\}  
    \end{equation}
    
    where $f_i$ denotes the DNN output representation of image $i$ and $w_{y_i}^c$ are the trainable centroids of class $y_i$. At inference time, the label is given to an image $i$ as follows: $\hat{y}_i = argmin_j||f_i-w_{y_j}^{c}||$. The total loss function for training the model is:
    
    \begin{equation}
        \mathcal{L}(x,y) = \mathcal{L}_{CE}(x,y)+ \mathcal{L}_{PC}(x,y)
    \end{equation}
    
    Where $\mathcal{L}_{CE}(.)$ is the cross-entropy loss. To obtain identical effect in intermediate layers, $\mathcal{L}_{PC}(.)$ is calculated using a branch $\mathcal{G}_{\phi}$, which maps the features to a lower dimension. So the final loss becomes:
    
    \begin{align*}
        \mathcal{L} & = \mathcal{L}_{CE} + \sum_{l}^{L}\mathcal{L}_{PC} \\
        & where \hspace{0.2cm} \mathcal{L}_{PC}^l = \mathcal{L}_{PC}(f^l,y)\\
        & s.t., \hspace{0.2cm} f^l = \mathcal{G}_{\phi}^l(F_{\theta}^l(x))
    \end{align*}

    \end{itemize}
    \subsubsection{Adversarial Detection}
    
    \begin{itemize}
        
    \item \textbf{MagNet}: \textit{MagNet} \cite{meng} is used for defending neural network classifiers against adversarial examples. MagNet consists of two components: 1) a \textit{detector} which rejects examples far from the decision boundary of different classes. 2) a \textit{reformer} which given an input $x$ strives to find
    $x{'}$ such that $x{'}$ is a close approximation of $x$ and also is close to the normal examples manifold(decision boundary).

    \textbf{Detector}: The detector is the function $d: \mathbb{S} \rightarrow {0,1}$ which detects whether the input is adversarial. Two kinds of detector were mentioned in the literature: 1) \textit{Detector based on Reconstruction Error} - In this an autoencoder is trained using the normal samples. The loss function minimized in this case is:
    
    \begin{equation*}
        L(\mathbb{X}_{train}) = \frac{1}{|\mathbb{X}_{train}|} \sum_{x \in \mathbb{X}_{train}} ||x-ae(x)||_2
    \end{equation*}
    
    $ae(x)$ denotes the reconstructed image obtained from the autoencoder $ae$. The reconstruction error on a test sample $x$ is:
    
    \begin{equation}
        E(x) = ||x-ae(x)||_p 
    \end{equation}
    
    where $E(x)$ is the $l_p$ distance between input image and its reconstructed image. If $E(x)$ is greater than some constant $\delta_p$ then the input is an adversarial example otherwise it is not. 2) \textit{Detector based on probability divergence} - Let $f(x)$ be the output of the last layer(softmax) of the neural network $f$ given an input $x$. Then in this case probability divergence of
    $f(x)$ and $f(ae(x))$ is evaluated using Jensen-Shannon Divergence:
    
    \begin{equation}
        JSD(P||Q) = \frac{1}{2} D_{KL}(P||M) + \frac{1}{2} D_{KL}(Q||M)
    \end{equation}
    
    where $D_{KL}(P||Q) = \sum_{i} P(i) log \frac{P(i)}{Q(i)}$ is the Kullback–Leibler Divergence and $M = \frac{1}{2}(P+Q)$. The Jensen–Shannon divergence is a method of measuring the similarity between two probability distributions $P$ and $Q$. $\\$
    $\\$
    \textbf{Reformer}: The reformer is the function $r:\mathbb{S} \rightarrow \mathbb{N}_t$ which tries to reconstruct the test input so that it is closer to the training data manifold. An ideal reformer: 1) should not change classification results of normal samples and 2) should change adversarial samples adequately such that the reformed samples are close to normal samples. Two kinds of reform strategies were proposed: 1) \textit{Noise based Reformer}: a naive reformer which adds random noise to the input sampled from a Gaussian distribution i.e $r(x) = clip(x+ \epsilon.y)$ where $y \sim \mathit{N}(y;0,I)$.
    2) \textit{Autoencoder based Reformer}: In this case the autoencoder trained for detector $ae(x)$ is used. For an input $x{'}$ the reconstruction obtained from autoencoder $ae(x{'})$ is fed to the target classifier as it is closer to the normal samples manifold.$\\$
    
     \item \textbf{Feature Squeezing}: Xu et al. \cite{xu} argues that feature input spaces are unnecessarily large. So he proposed a mechanism to reduce the degrees of freedom available to an adversary by squeezing out unnecessary input features. \textit{Feature Squeezing} is a simple adversarial example detection framework that compares the model's prediction on the original sample with its prediction after squeezing. If the difference is higher than a threshold value, then the sample is adversarial and is discarded.

    Two techniques have been proposed by the authors:1)\textit{Squeezing Color Bits}: Grayscale images comprise of 8-bit pixel values, whereas color images consist of 24-bit pixels. The bit depth can be considerably reduced without much loss of information, thereby reducing the search space for adversarial samples. This is done by using a binary filter with 0.5 as the cutoff. 

    2) \textit{Spatial Smoothing}: It is an image processing technique that is used to reduce noise. Its two variations have been mentioned:1) \textit{Local Smoothing}: Smoothes the value of a pixel in a particular window size by using nearby pixels. In this paper, the Median Smoothing was used. 2) \textit{ Non-Local Smoothing}: For a given image patch, non-local smoothing finds several similar patches in a larger area and then replaces the center patch with the average of those patches. \newline
    
    \item \textbf{Convex adversarial polytope}: In this paper, Wong et al. \cite{wong} proposed a method for obtaining \textit{provably robust} ReLU classifiers. For a ReLU based neural network $f$ where $f_{\theta} : \mathbb{R}^{|x|} \rightarrow \mathbb{R}^{|y|}$, we obtain an adversarial polytope ($\mathcal{Z}_{\epsilon}(x)$) which is the set of all final layer activations that can be obtained by perturbing the input with $\Delta$. Here, $\Delta$ is $l_{\infty}$ norm bounded by $\epsilon$:
    
    \begin{equation}
        \mathcal{Z}_{\epsilon}(x) = \{f_{\theta}(x + \Delta): ||\Delta||_{\infty} \leq \epsilon \}
    \end{equation}
    
    Now, a \textit{convex outer bound} on this adversarial polytope is created. If no point inside this \textit{outer bound} changes the class label of the input, then no label is changed in the true adversarial polytope as well. This bound is obtained by solving the following linear program:
    
    \begin{equation}
        z \geq 0,\hspace{1mm} z \geq \hat{z}, \hspace{1mm} -u\hat{z}+ (u-l)z \leq -ul
    \end{equation}
    
    Here, $u,l,\hat{z},z$ are the lower bound, upper bound, pre ReLU activation and ReLU activation respectively. Given a dataset $(x_i,y_i)_{i=1,2,...,N}$, instead of training the model using these data points, the model is trained using the farthest points (i.e with the highest loss) from $x_i$ which are $l_{\infty}$ by $\epsilon$:
    
    \begin{equation}
        minimize_{\theta} \sum_{i=1}^{N} max_{||\Delta||_{\infty} \leq \epsilon} L(f_{\theta}(x_i+\Delta),y_i)
    \end{equation}
    
    L is the loss function, and $\theta$ are the parameters of the feedforward ReLU neural network. 
    
    \end{itemize}
    
\subsection{Comparitive Analysis of Defense Techniques}

Similar to the attack section, the defense techniques have now been tested and compared on three datasets i.e., MNIST, CIFAR-10, and ImageNet. The results have been compared along the following dimensions: the attack model on which the defense technique was used, classification accuracy (CA), which is the accuracy obtained on clean images, attack success rate (ASR), which highlights the performance of the specific attack model and parameters used. CA and ASR have two subcategories as well: with defense and without defense which measure the above mentioned metrics with and without defense techniques. The results have been obtained from the respective research studies.

\begin{table}[h!]
  \centering
    \begin{adjustbox}{width = \textwidth, center}
    \begin{tabular}{|c|c|c|c|c|c|c|}
    \hline
    \textbf{Method} & \textbf{Attack Model} & \multicolumn{2}{c|}{\textbf{Classification Accuracy}} & \multicolumn{2}{c|}{\textbf{Attack Success Rate}} & \textbf{Specific Parameters Used} \\
          &       & \textbf{Without Defense} & \textbf{With Defense} & \textbf{Without Defense} & \textbf{With Defense} &  \\
    \hline
    Papernot \cite{papernot_defense} & JSMA \cite{jsma} & 99.51\% & 99.05\% & 95.89\% & 0.45\% & Temperature T = 100 \\
    \hline
    Hosseini \cite{hosseini} & FGSM \cite{FGSM}  & 99.35\% & 99.39\% & 90\%  & 45\%  & $\alpha = 0.5$, label Smoothing Parameter q = 0.9 \\
    \hline
    Meng \cite{meng}  & FGSM \cite{FGSM}  & 99.40\% & 99.10\% & 8.90\% & 0\%   & $L_{\infty}$ metric, $\epsilon = 0.010$, epochs = 100, lr = 0.001 \\
          & I-FGSM \cite{bim} & 99.40\% & 99.10\% & 28\%  & 0\%   & $L_{\infty}$ metric, $\epsilon = 0.010$, epochs = 100, lr = 0.001 \\
          & Deepfool \cite{deepfool} & 99.40\% & 99.10\% & 80.90\% & 0.60\% & $L_{\infty}$ metric, epochs = 100, lr = 0.001\\
          & C\&W \cite{cw}  & 99.40\% & 99.10\% & 100\% & 0.20\% & $L_{\infty}$ metric, epochs = 100, lr = 0.001 \\
    \hline
    Xu \cite{xu}    & FGSM \cite{FGSM}  & 99.43\% & 99.28\% & 46\%  & 8\%   & $L_{\infty}$, Bit Depth Method of 1 bit \\
          & C\&W \cite{cw}  & 99.43\% & 99.28\% & 100\% & 0\%   & $L_{\infty}$, Bit Depth Method of 1 bit \\
          & JSMA \cite{jsma}  & 99.43\% & 99.28\% & 71\%  & 18\%  & $L_{0}$ metric, Median Smoothing of $3x3$ \\
    \hline
    Buckman \cite{buckman} & FGSM \cite{FGSM}  & 99.30\% & 98.67\% & 100\% & 3.71\% & $l_{\infty}$ metric, $\epsilon = 0.3$, 40 steps for iterative attacks \\
          & PGD/LS-PGA \cite{PGD} & 99.30\% & 98.67\% & 100\% & 5.70\% & $l_{\infty}$ metric, $\epsilon = 0.3$, 40 steps for iterative attacks, $\xi = 0.01$, $\delta = 1.2$ \\
    \hline
    Samangouie \cite{samangouei} & FGSM \cite{FGSM}  & 99.70\% & -     & 71.84\% & 8.95\% & $\epsilon=0.3$, $L=200$, $R=10$\\
    \hline
    Kannan \cite{kannan} & PGD \cite{PGD}   & -     & 98.80\% & -     & 3.60\% & $l_{\infty}$ metric, $\epsilon=0.3$, lr = 0.045 \\
    \hline
    Wong \cite{wong}  & FGSM \cite{FGSM}  & 98.20\% & 94.18\% & 50.01\% & 3.93\% & $l_{\infty}$ metric, $\epsilon= 0.1$\\
          & PGD \cite{PGD}   & 98.20\% & 94.18\% & 81.68\% & 4.11\% & $l_{\infty}$ metric, $\epsilon= 0.1$ \\
    \hline
    Tramer \cite{tramer} & FGSM \cite{FGSM}  & -     & 99.30\% & 72.80\% & 14\%  & $l_{\infty}$ metric, $\epsilon=0.3$, lr = 0.045 \\
    \hline
    Madry \cite{madry2018} & FGSM \cite{FGSM}  & 99.20\% & 98.80\% & 58.10\% & 4.40\% & $l_{\infty}$ metric, $\epsilon=0.3$\\
          & PGD \cite{PGD}   & 99.20\% & 98.80\% & 74\%  & 10.70\% & $l_{\infty}$ metric, $\epsilon=0.3$, 40 iterations of PGD  \\
    \hline
    Mustafa \cite{mustafa} & FGSM \cite{FGSM}  & 98.71\% & 99.53\% & 77\%  & 21.70\% & epochs = 50, lr = 0.1, $\epsilon=0.3$ \\
          & C\&W \cite{cw}  & 98.71\% & 99.53\% & 79.10\% & 22.80\% & c = 10, 1000 iteration steps for C\&W \\
          & PGD \cite{PGD}   & 98.71\% & 99.53\% & 88.10\% & 30.50\% & 10 iteration steps for PGD with step size $\frac{\epsilon}{10}$ \\
    \bottomrule
    \end{tabular}%
    \end{adjustbox}
    \caption{Defense techniques against attacks on MNIST data set.}
    \label{defense_mnist}%
    
  \begin{adjustbox}{width = \textwidth, center}

  \begin{tabular}{|c|c|c|c|c|c|c|}
  \hline
    \textbf{Method} & \textbf{Attack Model} & \multicolumn{2}{c|}{\textbf{Classification Accuracy}} & \multicolumn{2}{c|}{\textbf{Attack Success Rate}} & \textbf{Specific Parameters Used} \\
          &       & \textbf{Without Defense} & \textbf{With Defense} & \textbf{Without Defense} & \textbf{With Defense} &  \\
    \hline
    Papernot \cite{papernot_defense} & JSMA \cite{jsma}     & 80.95\% & 81.34\% & 87.59\% & 5.11\% & Temperature T  = 100 \\
    \hline
    Meng \cite{meng}  & FGSM \cite{FGSM}  & 90.60\% & 86.80\% & 54\%  & 0.10\% & $L_{\infty}$ metric, $\epsilon = 0.050$, epochs = 400, lr = 0.001 \\
          & Iterative \cite{bim} & 90.60\% & 86.80\% & 88.90\% & 0.10\% & $L_{\infty}$ metric, $\epsilon = 0.025$, epochs = 400, lr = 0.001 \\
          & Deepfool \cite{deepfool} & 90.60\% & 86.80\% & 95.50\% & 6.60\% & $L_{\infty}$ metric, epochs = 400, lr = 0.001 \\
          & C\&W \cite{cw}  & 90.60\% & 86.80\% & 100\% & 17\%  & $L_{\infty}$ metric, epochs = 400, lr = 0.001 \\
    \hline
    Xu \cite{xu}    & FGSM \cite{FGSM}  & 94.84\% & 89.29\% & 85\%  & 62\%  & $L_{\infty}$ metric, Median Smoothing of $2x2$ \\
          & C\&W \cite{cw}  & 94.84\% & 89.29\% & 100\% & 16\%  & $L_{\infty}$ metric, Median Smoothing of $2x2$ \\
          & Deepfool \cite{deepfool} & 94.84\% & 89.29\% & 98\%  & 17\%  & $L_{2}$ metric, Median Smoothing of $2x2$ \\
          & JSMA \cite{jsma}  & 94.84\% & 89.29\% & 71\%  & 16\%  & $L_{0}$ metric, Median Smoothing of $2x2$ \\
    \hline
    Buckman \cite{buckman} & FGSM \cite{FGSM}  & 94.29\% & 87.67\% & 56.11\% & 19.04\% & $l_{\infty}$ metric, $\epsilon = 0.031$, 7 steps for iterative attacks\\
          & PGD/LS-PGA \cite{PGD} & 94.29\% & 87.67\% & 98.34\% & 20.84\% &  $l_{\infty}$ metric, $\epsilon = 0.031$, 7 steps for iterative attacks, $\xi = 0.01$, $\delta = 1.2$ \\
    \hline
    Dhillon \cite{dhillon} & FGSM \cite{FGSM}  & 89.80\% & -     & 30\%  & 22\%  & epochs = 150, Initial lr = 0.5 \\
          & I-FGSM \cite{bim} & 89.80\% & -     & 40\%  & 15\%  & $\lambda=1$, SAP-100 \\
    \hline
    Song \cite{song}  & FGSM \cite{FGSM}  & 92\%  & 88\%  & 89\%  & 76\%  & $l_{\infty}$ metric, $\epsilon_{attack}=2/8/16$, $\epsilon_{defend}=16$ \\
          & Deepfool \cite{deepfool} & 92\%  & 88\%  & 94\%  & 20\%  & $l_{\infty}$ metric, $\epsilon_{attack}=2/8/16$, $\epsilon_{defend}=16$ \\
          & C\&W \cite{cw}  & 92\%  & 88\%  & 100\% & 22\%  & $l_{\infty}$ metric, $\epsilon_{attack}=2/8/16$, $\epsilon_{defend}=16$ \\
    \hline
    Rakin \cite{rakin} & FGSM \cite{FGSM}  & 92.11\% & 84.78\% & 85.92\% & 45.96\% & $\epsilon=0.3/1$ and $\epsilon=8/255$ under $l_2$ and $l_{\infty}$ distance respectively \\
          & PGD \cite{PGD}   & 92.11\% & 84.78\% & 100\% & 54.17\% & Number of iteration steps is 7 for PGD \\
    \hline
    Cohen \cite{cohen} & Deepfool \cite{deepfool} & 90\%  & 78\%  & 80\%  & 35\%  & $l_2$ metric, $\alpha=0.001$, n = 100000 \\
    \hline
    Madry \cite{madry2018} & FGSM \cite{FGSM}  & 95.20\% & 90.30\% & 67.30\% & 4.90\% & $l_{\infty}$ metric, $\epsilon=8$ \\
          & PGD \cite{PGD}   & 95.20\% & 90.30\% & 96.50\% & 54.20\% & $l_{\infty}$ metric, $\epsilon=8$, 7 steps of size 2 for PGD\\
    \hline
    Mustafa \cite{mustafa} & FGSM \cite{FGSM}  & 90.80\% & 90.45\% & 61\%  & 14.50\% & epochs = 200, lr = 0.1, $\epsilon=0.03$ \\
          & C\&W \cite{cw}  & 90.80\% & 90.45\% & 68.20\% & 16.70\% & c = 0.1, 1000 iteration steps for C\&W \\
          & PGD \cite{PGD}   & 90.80\% & 90.45\% & 70.90\% & 23.60\% & 10 iteration steps for PGD with step size $\frac{\epsilon}{10}$ \\
    \bottomrule
    \end{tabular}%
    \end{adjustbox}

    \caption{Defense techniques against attacks on CIFAR-10 data set.}
    \label{defense_cifar}%
    
\end{table}

\begin{table}[h!]
\centering
\begin{adjustbox}{width = \textwidth,center}    
  
    \begin{tabular}{|c|c|c|c|c|c|c|}
    \hline
    \textbf{Method} & \textbf{Attack Model} & \multicolumn{2}{c|}{\textbf{Classification Accuracy}} & \multicolumn{2}{c|}{\textbf{Attack Success Rate}} & \textbf{Specific Parameters Used} \\
          &       & \textbf{Without Defense} & \textbf{With Defense} & \textbf{Without Defense} & \textbf{With Defense} &  \\
    \hline
    Xu \cite{xu}    & FGSM \cite{FGSM}  & 69.70\% & 62.10\% & 99\%  & 67\%  & $L_{\infty}$ metric, Median Smoothing of $3x3$ \\
          & C\&W \cite{cw}  & 69.70\% & 62.10\% & 99\%  & 23\%  & $L_{\infty}$ metric, Non Local Means Method (11-3-4) \\
          & DeepFool \cite{deepfool} & 69.70\% & 62.10\% & 89\%  & 28\%  & $L_{2}$ metric, Median Smoothing of $2x2$ \\
    \hline
    Guo \cite{guo_defense}   & FGSM \cite{FGSM}  & -     & 75.10\% & -     & 29.63\% & $L_{2}$ metric \\
          & I-FGSM \cite{bim} & -     & 75.10\% & -     & 28.48\% & Pixel dropout probability p = 0.5 \\
          & DeepFool \cite{deepfool} & -     & 75.10\% & -     & 28.53\% & Total Variance minimizer $\lambda_{TV} = 0.03$ \\
          & C\&W \cite{cw}  & -     & 75.10\% & -     & 29.50\% & Quilting Patch Size of $5x5$ \\
    \hline
    Xie \cite{xie_2018}   & FGSM \cite{FGSM}  & 100\% & 97.30\% & 67\%  & 47.60\% & $l_{\infty}$ metric, $\epsilon=10$ \\
          & DeepFool \cite{deepfool} & 100\% & 97.30\% & 100\% & 1.70\% & $l_{\infty}$ metric, $\epsilon=10$ \\
          & C\&W \cite{cw}  & 100\% & 97.30\% & 100\% & 3.10\% & $l_{\infty}$ metric, $\epsilon=10$ \\
    \hline
    Kannan \cite{kannan} & PGD \cite{PGD}   & -     & -     & -     & 72.10\% & $l_{\infty}$ metric, $\epsilon=0.3$, lr = 0.045 \\
    \hline
    Prakash \cite{prakash} & FGSM \cite{FGSM}  & 100\% & 98.90\% & 80\%  & 18.50\% & $l_2$ metric, $\epsilon=0.02-0.04$ \\
          & I-FGSM \cite{bim} & 100\% & 98.90\% & 85.90\% & 16.30\% & Coefficient for BayesShrink, $\sigma=0.04$ \\
          & DeepFool \cite{deepfool} & 100\% & 98.90\% & 73.70\% & 9.70\% & Window size for pixel deflection, r = 10 \\
          & JSMA \cite{jsma}  & 100\% & 98.90\% & 74.50\% & 3\%   & Number of pixel deflections, K = 100 \\
          & C\&W \cite{cw}  & 100\% & 98.90\% & 95.20\% & 2\%   & ResNet-50 Model \\
    \hline
    Liao \cite{liao}   & FGSM-White Box \cite{FGSM}  & 76.70\% & 75.30\% & 85.60\% & 63.30\% & epochs = 20 to 30, initial lr = 0.001, $l_{\infty}$ metric, $\epsilon=4/16$ \\
          & FGSM-Black Box \cite{FGSM} & 76.70\% & 75.30\% & 59\%  & 43.30\% & epochs = 20 to 30, initial lr = 0.001, $l_{\infty}$ metric, $\epsilon=4/16$ \\
    \hline
    Tramer \cite{tramer} & best(FGSM \cite{FGSM}, I-FGSM \cite{bim}, PGD \cite{PGD}) & 80.40\% & 79.80\% & 44.40\% & 27\%  & $l_{\infty}$ metric, $\epsilon=16$, lr = 0.045 \\
    \hline
    Xie \cite{xie_2019} & NIPS '17 CAAD \cite{nips_2017} & 78.91\% & 79.08\% & 100\% & 50.50\% & $l_{\infty}$ metric, $\epsilon=32$, $\alpha=1$ \\
    \hline
    Cohen \cite{cohen} & DeepFool \cite{deepfool} & 78\%  & 70\%  & 95\%  & 49\%  & $l_2$ metric, $\alpha=0.001$, n = 100000 \\
    \bottomrule
    \end{tabular}%
    \end{adjustbox}

    \caption{Defense techniques against attacks on ImageNet data set.}
    \label{defense_imagenet}%
    
\end{table}


Table \ref{defense_mnist} compares defenses for various attack models on MNIST dataset. Defenses proposed by Meng et al. \cite{meng} and Xu et al. \cite{xu} are the only ones for which \emph{Attack Success Rate} reduces to $0\%$. Distillation by Papernot et al. \cite{papernot_defense} is the defense method with the least reduction in classification accuracy. 

Table \ref{defense_cifar} shows similar comparison but instead on CIFAR-10 dataset with the following interesting observations:- 1) MagNet \cite{meng} demonstrates maximum robustness among the various defense techniques for FGSM \cite{FGSM} and I-FGSM \cite{bim} attacks. 2) The technique proposed by Mustafa et al. \cite{mustafa} demonstrates the highest generalizability in terms of classification accuracy. 

Table \ref{defense_imagenet} compares defense techniques on ImageNet dataset against various attacks like FGSM \cite{FGSM}, C\&W \cite{cw}, DeepFool \cite{deepfool}, PGD \cite{PGD} e.t.c. Average \emph{Attack Success Rates} on the ImageNet dataset are much higher as compared to MNIST and CIFAR-10 even when employing the same defense techniques. Pixel Deflection by Prakash et al. \cite{prakash} and Randomization by Xie at al. \cite{xie_2018} have the best generalizability in terms of classification. The models employed by them can get $100\%$ classification accuracy when not using the defense methods. Also, Pixel Deflection \cite{prakash} exhibits the highest robustness in comparison to other methods.

\section{Conclusion and Future Scope}
Our primary motive behind presenting this study is to encourage the community to push forward research in the black-box discipline. Analysis of the defense techniques discussed in this study reveals that the majority of these methods have been tested mostly on white-box adversarial attacks. To counter the security challenges put forward by the black-box adversaries, defense mechanisms should also be tested comprehensively on these models to enhance their robustness and applicability in a large number of scenarios. In addition, all attacks both white-box and black-box focus on perturbing all the features of the input image. We think that an interesting research direction would be the detection of non-robust features and exploiting only these features for misclassification.

\bibliographystyle{unsrt} 
\bibliography{references}  

\end{document}